\documentclass{article}

\usepackage{microtype}
\usepackage{graphicx}
\usepackage{booktabs} %
\usepackage[table]{xcolor}
\usepackage{makecell}
\usepackage{xspace}
\usepackage{siunitx}
\usepackage[abbreviations]{foreign}

\usepackage{caption}
\usepackage{subcaption}
\usepackage{multirow}
\usepackage{xcolor,colortbl}
\definecolor{lorautercol}{RGB}{230,245,255} %

\usepackage{hyperref}

\usepackage[preprint]{unknownconf}

\usepackage{amsmath}
\usepackage{amssymb}
\usepackage{mathtools}
\usepackage{amsthm}
\usepackage[inline]{enumitem}
\usepackage{pifont}

\usepackage[capitalize,noabbrev]{cleveref}
\theoremstyle{plain}

\theoremstyle{definition}

\theoremstyle{remark}
\usepackage{ifthen}
\newboolean{showcomments}
\setboolean{showcomments}{false}
\ifthenelse{\boolean{showcomments}}
{ \newcommand{\mynote}[3]{
		\fbox{\bfseries\sffamily\scriptsize#1}
		{\small$\blacktriangleright$\textsf{\emph{\color{#3}{#2}}}$\blacktriangleleft$}}
	\newcommand{\zzz}[1]{{\setlength{\fboxsep}{2pt}\fcolorbox{black}{yellow}{\textsf{\emph{#1}}}}\xspace}}
{ \newcommand{\mynote}[3]{}
	\newcommand{\zzz}[1]{}}

\definecolor{maroon}{rgb}{0.5, 0.0, 0.0}

\usepackage{acronym}
\acrodef{ML}{machine learning}
\acrodef{DL}{decentralized learning}
\acrodef{HbC}{honest-but-curious}
\acrodef{RMW}{random model walk}
\acrodef{D-PSGD}{decentralized parallel stochastic gradient descent}
\acrodef{FL}{federated learning}
\acrodef{SGD}{stochastic gradient descent}
\acrodef{IID}{independent and identically distributed}
\acrodef{non-IID}{non independent and identically distributed}
\acrodef{non-OOD}{in-domain}
\acrodef{OOD}{out-of-domain}
\acrodef{LLM}{large language model}
\acrodef{NLP}{natural language processing}
\acrodef{LoRA}{low-rank adaptation}
\acrodef{PEFT}{parameter-efficient fine-tuning}
\acrodef{MOE}[MoE]{Mixture-of-Experts}
\acrodef{NLU}{natural language understanding}
\acrodef{NLG}{natural language generation} %
\newcommand{\sys}{\textsc{LoRAuter}\xspace}

\newcommand{\arrow}{\textsc{Arrow}\xspace}
\newcommand{\mole}{\textsc{MoLE}\xspace}
\newcommand{\lorahub}{\textsc{LoraHub}\xspace}
\newcommand{\loraretriever}{\textsc{LoraRetriever}\xspace}
\newcommand{\adaptersoup}{\textsc{AdapterSoup}\xspace}

\newcommand{\R}{\mathcal{R}\xspace}
\newcommand{\T}{\mathcal{T}\xspace}
\newcommand{\M}{\mathcal{M}\xspace}
\newcommand{\e}{\mathbf{e}\xspace}
\allowdisplaybreaks

\graphicspath{ {figures/} }
\usepackage{tikz}
\usepackage{pgfplots}
\usepackage{pgfplotstable}
\usepackage[eulergreek]{sansmath}
\usepackage{comment}
\newcolumntype{L}[1]{>{\raggedright\let\newline\\\arraybackslash\hspace{0pt}}m{#1}}
\newcolumntype{C}[1]{>{\centering\let\newline\\\arraybackslash\hspace{0pt}}m{#1}}
\newcolumntype{R}[1]{>{\raggedleft\let\newline\\\arraybackslash\hspace{0pt}}m{#1}}

\pgfplotsset{compat=newest}
\usepgfplotslibrary{external,units,colorbrewer,groupplots,fillbetween}
\tikzsetexternalprefix{figures/}
\tikzset{external/mode=list and make}
\usetikzlibrary{patterns}
\usetikzlibrary{shapes.geometric, arrows, positioning}

\def\overleafhome{/tmp}
\newcommand{\inputplot}[2]{%
	\ifx\homepath\overleafhome%
	\IfBeginWith{#1}{plots}{\includegraphics{main-figure#2.pdf}}{#1}%
	\else%
	{\sffamily\scriptsize\input{#1}}
	\fi
}

\newcommand{\newgroupwidth}[2]%
{\expandafter\xdef\csname groupwidth#1\endcsname{#2}}

\newcounter{groupwidth}
\newsavebox{\groupwidthbox}
\makeatletter
{\edef\groupnumber{#1}%
	\stepcounter{groupwidth}%
	\@ifundefined{groupwidth\thegroupwidth}{\pgfmathsetlengthmacro{\mywidth}{\linewidth/\groupnumber}}%
	{\expandafter\let\expandafter\mywidth\csname groupwidth\thegroupwidth\endcsname}%
	\begin{lrbox}{\groupwidthbox}%
		\tikzset{/pgfplots/width={\mywidth}}%
		\ignorespaces}%
	{\end{lrbox}%
	\usebox\groupwidthbox
	\pgfmathsetlengthmacro{\mywidth}{\mywidth + (\linewidth - \wd\groupwidthbox)/\groupnumber}
	\immediate\write\@auxout{\string\newgroupwidth{\thegroupwidth}{\mywidth}}}
\makeatother

\makeatletter
\AtBeginDocument{%
	\def\ltx@label#1{\cref@label{#1}}%

	\def\label@in@display@noarg#1{\cref@old@label@in@display{#1}}%

	\def\label@in@mmeasure@noarg#1{%
		\begingroup
		\measuring@false
		\cref@old@label@in@display{#1}%
		\endgroup
	}%
}
\makeatother

\usepackage[textsize=tiny]{todonotes}
\usepackage{placeins}

\icmltitlerunning{Effective LoRA Adapter Routing using Task Representations}

\begin{document}

\twocolumn[
\icmltitle{Effective LoRA Adapter Routing using Task Representations}

\icmlsetsymbol{equal}{*}

\begin{icmlauthorlist}
\icmlauthor{Akash Dhasade}{EPFL}
\icmlauthor{Anne-Marie Kermarrec}{EPFL}
\icmlauthor{Igor Pavlovic}{EPFL}
\icmlauthor{Diana Petrescu}{EPFL}
\icmlauthor{Rafael Pires}{EPFL}
\icmlauthor{Mathis Randl}{EPFL}
\icmlauthor{Martijn de Vos}{EPFL}
\end{icmlauthorlist}

\icmlaffiliation{EPFL}{EPFL, Lausanne, Switzerland}

\icmlcorrespondingauthor{Igor Pavlovic}{igor.pavlovic@epfl.ch}

\icmlkeywords{Machine Learning, ICML}

\vskip 0.3in
]

\printAffiliationsAndNotice{\icmlEqualContribution} %

\begin{abstract}

\Ac{LoRA} enables parameter-efficient specialization of \acp{LLM} through modular adapters, resulting in rapidly growing public adapter pools spanning diverse tasks.
Effectively using these adapters requires routing: selecting and composing the appropriate adapters for a query.
We introduce \sys, a novel routing framework that selects and composes \ac{LoRA} adapters using \emph{task representations} rather than adapter characteristics.
Unlike existing approaches that map queries directly to adapters, \sys routes queries via task embeddings derived from small validation sets and does not require adapter training data.
By operating at the task level, \sys achieves efficient routing that scales with the number of tasks rather than the number of adapters.
Experiments across multiple tasks show that \sys consistently outperforms baseline routing approaches, matching Oracle performance (\num{101.2}\%) when task-aligned adapters exist and achieving state-of-the-art results on unseen tasks (+\num{5.2} points).
We further demonstrate the robustness of \sys to very large, noisy adapter pools by scaling it to over \num{1500} adapters.

\end{abstract}
\section{Introduction}
\label{sec:introduction}

Modern \acfp{LLM} such as GPT~\cite{openai2025gpt5}, Gemini~\cite{team2023gemini} and Llama 4~\cite{touvron2023llama} have driven remarkable progress in \ac{NLP} and beyond.
However, specializing such massive monolithic models for each downstream task (\eg question answering or translation) remains computationally expensive and memory-intensive.
To address these challenges, \textit{\ac{PEFT}} methods \cite{lester2021prompttuning,liu2022p,hu2022lora} have emerged as effective alternatives, enabling adaptation by updating only a small subset of parameters or introducing lightweight, trainable modules.
Among them, \emph{\acf{LoRA}}~\cite{hu2022lora} has gained particular prominence due to its simplicity, modularity, and strong performance.
\Ac{LoRA} injects small, low-rank matrices into frozen model weights, allowing efficient fine-tuning while preserving the general knowledge of the base model.
Its modular design has bootstrapped large, public adapter pools, \eg, there are over \num{2300} adapters available just for the \textsc{meta-llama/Llama-2-7b-hf} model on \textsc{HuggingFace} for a wide variety of tasks\footnote{See \url{https://huggingface.co/models?other=base_model:adapter:meta-llama/Llama-2-7b-hf}}.

Effectively using these adapter pools during inference introduces a critical bottleneck: \textit{adapter routing}.
Given a user query, how do we efficiently identify and select an appropriate subset of adapters?
Existing approaches for \ac{LoRA} adapter routing formulate this as a granular retrieval problem, attempting to map each individual query directly to a small number of adapters, typically between one and four. %
However, this strategy faces several limitations.
First, existing methods either rely on a \emph{static pool of adapters}~\cite{huang2024lorahub,wu2024mixture} or require access to the original \emph{adapter training data} to construct a searchable index~\cite{zhao2024loraretriever,chronopoulou2023adaptersoup}, conditions that rarely hold for dynamic, public adapter pools where training data is proprietary or withheld for privacy reasons.
Second, methods that overcome this issue by directly routing based on adapter parameter values~\cite{ostropenko2024arrow} incur routing overheads that scale linearly with the number of \emph{adapters and layers}, making them expensive for large libraries.
Finally, formulating routing as a direct matching problem between queries and adapters fails to exploit the inherent structure of the problem: adapters are not granular experts for individual queries, but specialized modules designed to solve specific \emph{tasks}.

We propose a shift in perspective: instead of mapping queries directly to adapters, we identify \emph{task representations} as the necessary intermediary for \ac{LoRA} adapter routing.
We argue that routing should operate at the task level, mapping queries to semantic task clusters (\eg, translation or logic) rather than directly to adapters.
Based on this insight, we introduce \sys, a \emph{training-free} framework for \ac{LoRA} adapter routing that operates in a black-box setting without access to adapter training data.
\sys organizes an unstructured adapter pool into a task-indexed catalog using lightweight validation sets, retrieves the most relevant tasks for a given query, and composes the corresponding adapters in an input-aware manner.

Concretely, \sys first constructs a database of representative tasks using small, publicly available validation sets.
For each task, it identifies the most suitable adapter from the pool using efficient search strategies, such as Successive Halving~\citep{jamieson2016nonstochastic}, which substantially reduces evaluation cost.
At inference time, incoming queries are embedded and matched against task representations, allowing \sys to retrieve the top-$K$ relevant tasks and their associated adapters.
These adapters are then composed using a weighted output-space fusion mechanism that reflects input–task similarity, enabling effective generalization to both in-domain and out-of-domain queries.

We implement \sys and evaluate our framework on a mixed-task benchmark comprising diverse tasks and \num{48} adapters~\cite{zhao2024loraretriever}.
Our results demonstrate that \sys achieves \num{101.2}\% of the performance of an Oracle task-aligned adapter in in-domain settings, effectively outperforming the upper bound of selecting the perfect adapter for a query.
In out-of-domain scenarios, where the model must generalize to unseen tasks, \sys outperforms the strongest baseline, \loraretriever~\cite{zhao2024loraretriever}, by \num{5.2} percentage points.
We then scale our system to a pool of 1500+ adapters fetched \textit{from the wild}, and show that \sys achieves competitive performance to the original \num{48} adapters on the same mixed-task benchmark.
Thus, \sys can extract useful routing signals from large, noisy adapter collections, offering a scalable and robust solution for open-ended \ac{LoRA} serving.

Our main contributions are:

\begin{itemize}[leftmargin=*]
    \item \textbf{Training-free, black-box routing:} We introduce a \ac{LoRA} routing framework that requires neither adapter training data nor router model training, enabling black-box use of large and heterogeneous adapter pools.
    \item \textbf{Efficiency:} \sys scales efficiently with the number of \emph{tasks}, which is typically much smaller than the number of available adapters. We also employ a search strategy (Successive Halving) to quickly identify optimal adapters for representative tasks, reducing computational overhead by over $2\times$ compared to full adapter evaluation.
    \item \textbf{Extensive evaluation:} We conduct comprehensive evaluations across large adapter pools, multiple model sizes, and ablation settings, demonstrating the effectiveness of \sys.
\end{itemize}
\section{Background and problem description}
\label{sec:preliminaries}

\begin{table*}[t]
	\vskip -0.1in
	\caption{A qualitative comparison of recently proposed model routing approaches. $N, T$ and $L$ refer to the total adapters in the pool, number of representative tasks, and number of layers in the model. We typically have $T < N$.
	}
	\label{tab:method-comparison}
	\centering
	\small
		\begin{tabular}{lcccc}
			\toprule
			\textbf{Method} & \textbf{\makecell{Training data\\not required}} & \textbf{\makecell{Training \\ free}} &  \textbf{\makecell{Adapter Selection \\Overhead}} \\
			\midrule
			\mole$_{\color{gray}\text{ [ICLR24]}}$ \cite{wu2024mixture} & $\surd$ & $\times$ &  $O(NL)$ \\
			\lorahub$_{\color{gray}\text{ [COLM24]}}$ \cite{huang2024lorahub}  & $\surd$ & $\times$ &  $O(1)$ \\
			\adaptersoup$_{\color{gray}\text{ [FindingsEACL23]}}$~\cite{chronopoulou2023adaptersoup} & $\times$ & $\surd$ &  $O(1)$ \\
			\loraretriever$_{\color{gray}\text{ [FindingsACL24]}}$ \cite{zhao2024loraretriever}   & $\times$ & $\surd$ & $O(N)$ \\
			\arrow$_{\color{gray}\text{[ICML24]}}$ \cite{ostropenko2024arrow} & $\surd$ & $\surd$ &  $O(NL)$ \\
			\midrule
			\textbf{\sys (Ours)}             & $\surd$ & $\surd$ & $O(T)$ \\
			\bottomrule
		\end{tabular}
\end{table*}

\subsection{\Acf{LoRA}}
\label{subsec:lora}
\Acf{LoRA}~\citep{hu2022lora} is a \acf{PEFT} technique that enables adapting \acp{LLM} to excel on downstream tasks with minimal additional parameters.
Instead of updating all model weights, \ac{LoRA} trains a pair of low-rank matrices for selected weight layers, while keeping the original weights frozen.
Formally, for a weight matrix $W \in \mathbb{R}^{m \times d}$, \ac{LoRA} learns two smaller matrices $A \in \mathbb{R}^{r \times d}$ and $B \in \mathbb{R}^{m \times r}$ such that the effective parameter update is given by $\Delta W = B A$, where $r \ll \min(m, d)$ denotes the adaptation rank.
During inference, the forward computation becomes $h' = (W + \Delta W) x = Wx + B A x$, allowing efficient adaptation while keeping the number of new parameters small.
\ac{LoRA} achieves competitive performance compared to full fine-tuning but substantially reduces storage and computation costs.

\subsection{Composing \ac{LoRA} adapters}
\label{sec:lorafusion}
Each trained \ac{LoRA} adapter $(B_i, A_i)$ captures task-specific adaptations of the base model.
Combining multiple adapters can therefore integrate complementary skills and knowledge across tasks.
Early fusion strategies, \eg, parameter-space interpolation~\citep{zhang2023composing}, merge adapters linearly:
\begin{equation}
h' = Wx + (\lambda \Delta W_1 + (1 - \lambda) \Delta W_2) x,
\end{equation}
where $\lambda \in [0,1]$ controls the relative contribution of each adapter.
However, this approach requires manual tuning of $\lambda$ for each task.
More advanced approaches, such as \lorahub~\citep{huang2024lorahub}, extend this idea by learning adaptive fusion weights rather than relying on manually tuned coefficients.
Given a set of \ac{LoRA} adapters, the fused representation is computed as:
\begin{equation}
h' = Wx + \left(\sum_i w_i B_i\right)\left(\sum_i w_i A_i\right)x,
\end{equation}
where $w_i$ are learnable weights optimized in a few-shot manner. Although \lorahub effectively automates the fusion process, the learned weights still remain task-specific, requiring separate optimization for each new task and incurring additional computational overhead.

\subsection{Problem formulation}
We formalize the problem of \ac{LoRA} adapter routing.
We consider a pool of $N$ \ac{LoRA} adapters, $\Phi = \{\phi_1, \phi_2, \ldots, \phi_N\}$, available for routing, where each adapter $\phi_i = (B_i, A_i)$.
In practice, an adapter is not made of a single pair of matrices, but of a pair of matrices for every layer of the model that benefits from the adapter~\cite{hu2022lora}.
To simplify notation, we consider these adapter layers to be implicit.

Let $\mathcal{T}^{tr} = \{t_1^\text{tr}, t_2^{tr},\ldots,t_M^{tr}\}$ denote the collective training task set.
Each adapter $\phi_i$ is associated with a task $t_j^{tr} \in \T^{tr}$.
We allow $ M \leq N$, since a single task may have multiple adapters trained for it (\eg with different ranks or training procedures)\footnote{\url{https://huggingface.co/tasks} provides examples where multiple models are trained for the same task.}.
In the case when $M = N$, tasks and adapters have one-to-one correspondence.
Contrary to \loraretriever~\cite{zhao2024loraretriever}, we assume that no data from the original training set is available for routing.
However, in line with \citet{zhao2024loraretriever}, we consider user queries at test-time which can span a wide variety of underlying tasks \textit{without} an explicit task label.
In particular, a query $x$ can be both \ac{non-OOD} \ie $\texttt{task}(x) \in \mathcal{T}^{tr}$ or \ac{OOD} \ie $\texttt{task}(x) \notin \mathcal{T}^{tr}$ where \texttt{task}(.) indicates the ground-truth task label of the query.

The goal of routing is to efficiently \textit{select} and \textit{compose} adapters from the pool to respond to a given user query, generalizing to unseen tasks or domains.
In this sense, we seek an open-ended multi-\ac{LoRA} serving framework, capable of dynamically adapting to heterogeneous requests as new \ac{LoRA} adapters are added to the pool, extending the vision outlined by \citet{zhao2024loraretriever}.
Under this scenario, given a query $x$ without its task label, the serving process can be expressed as: $y = F(\R_x, x, W)$
where $W$ refers to the weights of base \ac{LLM}, $\R_x$ is the output of the selection or retrieval mechanism for query $x$ and $F$ is the composition function.
For instance, $\R_x$ can correspond to the top-$K$ \ac{LoRA} adapters retrieved for answering $x$ and $F$ can correspond to one of the aforementioned fusion strategies.

\subsection{Shortcomings of existing approaches}

Table~\ref{tab:method-comparison} summarizes existing \ac{LoRA} adapter routing approaches along key dimensions, including whether they \emph{(i)} require access to adapter training data, \emph{(ii)} involve additional training for routing or composition, and \emph{(iii)} incur inference-time overhead that scales with the number of adapters~$(N)$ and model layers~$(L)$.
Our objective is a routing mechanism that is \emph{training-free}, operates in a \emph{black-box} setting without access to adapter training data, and remains computationally efficient as the adapter pool grows.

Many prior methods violate one or more of these desiderata.
\ac{MOE}-style approaches such as \mole~\cite{wu2024mixture} rely on learned gating networks trained with task-specific data, which must be retrained for each task and limits applicability in open-ended settings.
Similarly, \lorahub~\cite{huang2024lorahub} learns adapter composition weights using validation data, requiring task-specific supervision at deployment time.

Other approaches avoid training but still depend on adapter training data.
\adaptersoup~\cite{chronopoulou2023adaptersoup} and \loraretriever~\cite{zhao2024loraretriever} embed adapters using training data and perform routing by matching queries to these representations.
While effective, such methods are unsuitable for public or proprietary adapter pools where training data is unavailable.
In contrast, spectral routing methods such as \arrow~\cite{ostropenko2024arrow} eliminate both training data and learned routing components by operating directly on adapter parameters, but incur higher adapter selection overheads due to layer-wise routing.

Beyond data and training requirements, existing methods also differ substantially in inference-time selection overhead.
Methods like \lorahub and \adaptersoup produce a single composite adapter per task and incur constant selection cost, while \mole and \arrow perform routing at every layer, leading to $O(NL)$ complexity.
\loraretriever avoids layer-wise routing but still compares each query against all adapters, resulting in $O(N)$ complexity.

In contrast, \sys operates at the level of \emph{tasks} rather than individual adapters.
By routing queries to task representations, \sys reduces selection complexity to $O(T)$, where $T$ denotes the number of representative tasks (\cref{sec:task_database}) and typically satisfies $T < N$.
This design enables scalable, training-free routing that remains efficient as the adapter pool grows.

\section{Design of \sys}
\label{sec:method}

\begin{figure*}[ht]
\centering
\includegraphics[width=\textwidth]{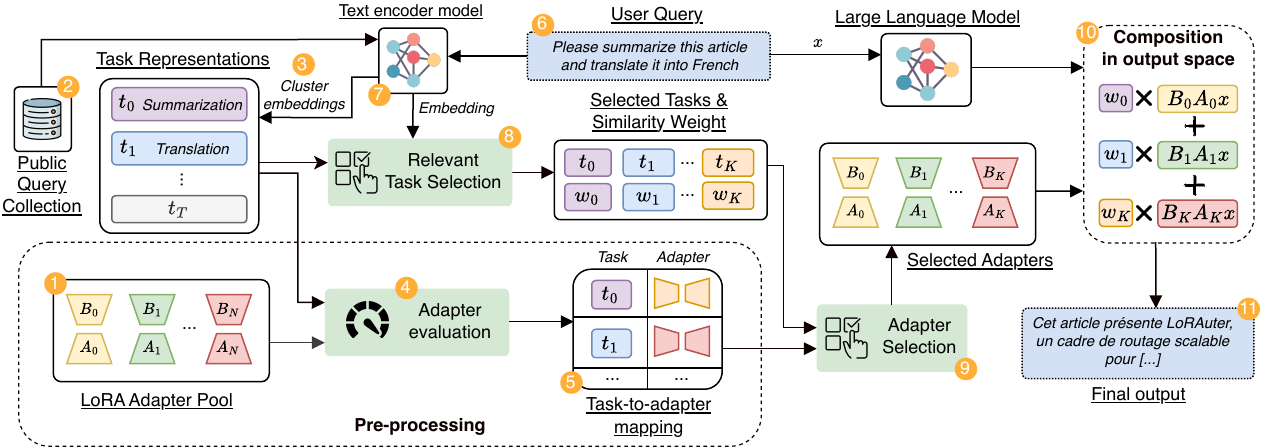}
\caption{The LoRAuter framework.}
\label{fig:lorauter-framework}
\end{figure*}

We now present the design of \sys.
The central idea of \sys is to perform \ac{LoRA} adapter selection and composition at the level of \emph{tasks}, rather than directly reasoning over individual adapters.
To this end, \sys is organized into four core components: \emph{(1)} a database of relevant tasks, \emph{(2)} task-to-adapter pairing, \emph{(3)} query-based task retrieval, and \emph{(4)} adapter composition.
The overall workflow of \sys is illustrated in \Cref{fig:lorauter-framework}.

\subsection{Database of relevant tasks ({\color{orange}\ding{203}})}
\label{sec:task_database}
\sys begins by constructing a representative task database $\T^{rep} = \{t_1, t_2, \ldots, t_T\}$ from publicly available task queries.
This task database serves as a repository of $T$ diverse tasks representing the range of query types the system may encounter.
Each task $t_i$ is accompanied by a small validation set $D_i$ containing labeled samples, which are lightweight and readily available in practice today.
Note that $T$ can be typically much lower than $N$, the total number of adapters in the pool.

\subsection{Task to adapter pairing ({\color{orange}\ding{204}, \ding{205}, \ding{206}})}
Once the task database has been constructed, the next step is to determine the most suitable LoRA adapter for each task.
To build this mapping $\M : \T^{rep} \rightarrow \Phi$, we evaluate every adapter on the small validation set associated with a given task and assign the adapter that yields the highest score according to that task’s predefined evaluation metric (\eg, ROUGE for generation tasks or BLEU for translation).
This procedure also allows flexible system updates, where new adapters or tasks can be added or removed, and the target evaluation metric for any task can be modified with minimal overhead.

Since both the number of tasks and available adapters can be substantial, and most adapters are only effective for a limited subset of tasks, exhaustive evaluation quickly becomes computationally prohibitive.
To address this, we employ tournament-based search strategies, such as \textit{Successive Halving}~\citep{jamieson2016nonstochastic} (SH), which iteratively eliminate underperforming adapters while performing additional evaluations on more promising candidates.
This efficiently identifies the best adapter for each task while substantially reducing computational overhead.
Furthermore, SH scales to hundreds of adapters by parallelizing evaluation rounds across GPUs with minimal coordination.

Alternative methods, such as \lorahub~\citep{huang2024lorahub}, can also be used to determine optimal adapters by learning fusion weights. However, these techniques exhibit limited scalability, as the training process for adapter fusion is typically more computationally expensive than adapter evaluation.
Notably, our framework can easily incorporate such training-based methods by treating the fused adapter produced by \lorahub (or similar approaches) as an additional candidate within the adapter pool.
If the fused adapter surpasses the current best adapter for a task, it becomes the new selected adapter.
We leave the integration and evaluation of these methods as directions for future work.

\subsection{Query-based task retrieval ({\color{orange}\ding{207}, \ding{208}, \ding{209}, \ding{210}})}

\sys identifies the most relevant task for a given query using task representations.
These representations are constructed in a one-time offline manner by employing a sentence-embedding model $E$.
Specifically, the model $E$ encodes a subset of the tasks' validation queries to obtain the task representation as follows.

For each task $t_i$, we construct a fixed embedding by encoding and aggregating a small subset of its validation queries.
Let $\{v_j\}_{j=1}^{m} \subset D_i$ be a randomly sampled subset with $m \ll |D_i|$.
Given an embedding instruction $I$ and a validation query $v$, the encoder $E$ produces an embedding by encoding the concatenation $I \oplus v$.
The task representation $\mathbf{e}_i$ is then computed as:
\[
\mathbf{e}_i = \frac{1}{m} \sum_{j=1}^{m} E(I \oplus v_j).
\]

Similar to \cite{zhao2024loraretriever}, we set $I$ to the following instruction: \textit{``Represent the sentence for similar task retrieval''}, allowing the embedding to capture task similarity.

At inference time, we similarly encode an arriving query $x$ to obtain an embedding $\e_x$.
This embedding is then compared against the precomputed task embeddings $\{\e_i\}_{i=1}^{T}$ using the \textit{cosine similarity} metric:
\begin{equation}
s_i = \frac{\mathbf{e}_x \cdot \mathbf{e}_i}{\|\mathbf{e}_x\| \, \|\mathbf{e}_i\|},
\end{equation}
where $s_i$ denotes the similarity between the input and the task $t_i$.
We then retrieve the set of top-$K$ most similar tasks based on their scores $s_i$: $\T_x = \{t_i \mid s_i \text{ is in top-}K\}$. To obtain a probabilistic interpretation of the input-task affinities, we normalize these scores using the \textit{softmax} function with a temperature parameter $\tau$.
The resulting probability $p_i$ reflects the likelihood that the input belongs to task $t_i$.
The sentence encoder used in this paper is trained using the \textit{Supervised Contrastive (SupCon)} loss~\citep{khosla2020supervised}, which minimizes a temperature-scaled cross-entropy objective over cosine similarities.
Thus, applying a softmax function to cosine similarities yields a probabilistic interpretation consistent with the encoder’s training objective.
\subsection{Adapter composition ({\color{orange}\ding{211}})}

After retrieving the top-$K$ tasks $\T_x$ for a given query $x$ along with corresponding probabilities $\{p_i\}_{i=1}^{K}$, we use the precomputed task-to-adapter mapping $\M$ to identify the adapter $\phi \in \Phi$ associated with each task $t_i$.
We define $\R_x = \{ (p_i, \M[t_i]) \mid t_i \in \T_x\}$ as the output of selection step, containing relevant adapters coupled with their matching likelihoods.
The goal is now to compose these $K$ adapters to produce the final output, leveraging their complementary strengths.

While prior methods \cite{zhao2024loraretriever,chronopoulou2023adaptersoup} treat all retrieved tasks as equally relevant by using uniform averaging for composition, we use an input-aware fusion mechanism that dynamically adjusts the contribution of each adapter based on the input-task similarity.
Specifically, the fused adapter outputs $h'$ are obtained through a weighted combination of the top-$K$ adapters in the \textit{output space}:
\begin{equation}
h' = W x + \sum_{(w_i, \phi_i) \in \R_x} w_i \, B_i A_i x,
\label{eq:adapter_fusion}
\end{equation}
where $\sum_{i=1}^{K} w_i = 1$ where the weights $w_i$ correspond to probabilities $p_i$.
This formulation adaptively integrates information from task-relevant adapters, allowing the model to emphasize those most aligned with the input while minimizing interference from unrelated tasks.
Furthermore, composition in the output space yields better performance than composition in the parameter space~\cite{zhao2024loraretriever}, as \ac{LoRA} adapters often exhibit limited alignment, which hinders effective parameter merging~\cite{stoica2025model}.

\section{Evaluation}

We next evaluate the effectiveness of the \sys framework using representative benchmarks and against four state-of-the-art routing baselines.
Our evaluation proceeds as follows: we first report main performance results under non-OOD, OOD, and semi-OOD settings (see~\Cref{fig:ood_protocol}, Appendix~\ref{appendix:evaluation_details}), and study robustness when scaling to large adapter pools (\S\ref{sec:main_results});
we then conduct two ablation studies to assess the contributions of retrieval and composition (\S\ref{sec:ablation});
we examine performance when task labels are replaced with data-driven pseudo-tasks obtained via clustering (\S\ref{sec:clusters});
and finally, we evaluate the efficiency of adapter selection using Successive Halving (\S\ref{sec:adapter_eval}).

\subsection{Experimental setup}

\textbf{Dataset.}
We adopt a curated subset of 48 individual tasks from \textsc{Flanv2} \citep{wei2022finetuned}, covering a broad range of Natural Language Understanding (NLU) and Natural Language Generation (NLG) settings.
We note that the same set of tasks is used in \citet{zhao2024loraretriever}, enabling a direct comparison between \sys and \loraretriever.

\textbf{Adapters.}
We use LLaMA~2-{7B,13B}~\citep{touvron2023llama2openfoundation} as the base model, following prior work and leveraging its large ecosystem of publicly available \ac{LoRA} adapters.
Each task from the above dataset is treated independently and used to train a dedicated \ac{LoRA} adapter, resulting in a pool of 48 task-specific adapters.
Each \ac{LoRA} adapter is trained using the Alpaca instruction-following format \citep{taori2023stanford}.
The rank $r$ and scaling hyperparameter $\alpha$  are fixed to 6 and 12, respectively, matching the configuration in the \loraretriever baseline.
Following standard \ac{LoRA} practice, the learned low-rank weight update is scaled by $\alpha/r$.
To further ensure comparability, we also adopt the same text encoder used in \citet{zhao2024loraretriever} for generating task embeddings and input representations.
Lastly, we also assess \sys on a public adapter pool fetched from \textsc{HuggingFace}, comprising 1500+ adapters.

\textbf{Representative tasks.}
In the \ac{non-OOD} routing setting, $\T^{rep} = \T^{tr}$ as is expected.
While in the \ac{OOD} routing setting, we set $\T^{rep} = \T^{tr} \backslash \{ \texttt{task}(x) \}$ \ie we explicitly remove the ground truth task corresponding to the query $x$ being evaluated, and the adapter trained on this task.
In either setting, \sys does not use any training data, but only the validation sets corresponding to tasks in $\T^{rep}$, with at most 200 validation samples per task.
Despite this limited validation set, we show that it is sufficient to reliably identify the best-performing adapter from the pool.
For the final evaluation, we use the same test set as \citet{zhao2024loraretriever}, consisting of 50 held-out test samples for each of the 48 tasks.

\textbf{Baselines.}
We compare \sys against the following approaches: \lorahub (\citealp{huang2024lorahub}), a universal fusion method that learns global mixture weights across; \loraretriever (\citealp{zhao2024loraretriever}), a retrieval-based method that ranks adapters using input to training data similarity scores and performs fusion over the top-$K$ most similar adapters using uniform weights (mixture); ARROW (\citealp{ostropenko2024arrow}), a spectral routing approach that applies SVD to adapter representations and selects adapters based on spectral affinity with the input; and SpectR (Spectral Routing) (\citealp{fleshman2024spectr}), an eigenspace-based spectral routing variant that projects inputs and adapters into aligned eigenspaces for similarity-based selection.
As an oracle, we report the performance achieved when directly using the adapter trained on the same task.
Note that the oracle is fixed, and remains independent of the \ac{OOD} or \ac{non-OOD} setting.

\textbf{Metrics.}
Following \citet{wei2022finetuned} and \citet{zhao2024loraretriever}, task performance is measured using task-specific metrics: Exact Match (EM) accuracy for NLU tasks, BLEU for translation tasks, and ROUGE-1, ROUGE-2, and ROUGE-L for structure-to-text generation tasks.
To aggregate performance across tasks with different metrics, we normalize each score by dividing it by the oracle performance for the corresponding task.
We then average the normalized scores across tasks and report the resulting metric.

\begin{figure}[t]
    \centering
    \includegraphics{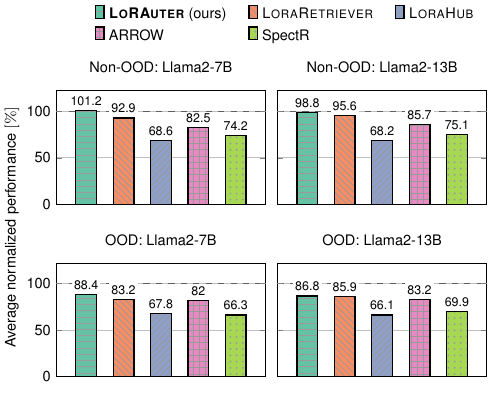}
    \caption{Average normalized performance of different routing methods on non-OOD and OOD tasks for both models.}
    \label{fig:normalized_acc}
\end{figure}

\subsection{Main results}
\label{sec:main_results}

\textbf{OOD and non-OOD settings.}
\Cref{fig:normalized_acc} presents the results for both the models (Llama2-7B \& Llama2-13B) and settings.
The corresponding detailed numerical results are included in \Cref{tab:original_adapter_results_with_full_fusion} (Appendix~\ref{app:per_task_group_results}).
Our proposed \sys achieves the strongest overall performance across each configuration.
Under \ac{non-OOD} setting and Llama2-7B, it attains an average normalized score of 101.2\%, slightly surpassing the oracle (100\%).
This shows that uncertainty-aware fusion of complementary adapters from related tasks can sometimes even outperform the adapter trained for a given task.
Under the \ac{OOD} setting and Llama2-7B, \sys reaches 88.4\%, outperforming \loraretriever, the best baseline by +5.2 points.
The same pattern holds for Llama2-13B.
\sys achieves 98.8\% on \ac{non-OOD} evaluation, again closely matching the oracle, and 86.8\% in the \ac{OOD} setting.
Spectral routing methods (\arrow and SpectR) achieve relatively lower performance as they directly route based on parameter values, which may not carry sufficient routing signal.
As with the 7B model, \sys outperforms baselines in each setting, demonstrating that the advantages of uncertainty-aware composition persist even on larger models.

\begin{figure}[t]
    \centering
    \includegraphics{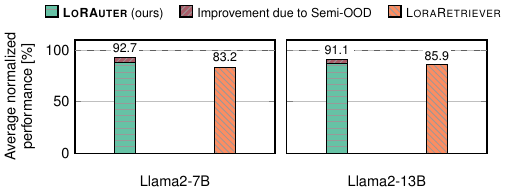}
    \caption{Performance in the semi-OOD setting where validation data is available but the task-specific adapter is removed.}
    \label{fig:lrvsla}
\end{figure}

\textbf{Semi-OOD setting.}
We introduce a \emph{semi-OOD} setting that lies between the non-OOD and OOD regimes and reflects realistic deployment scenarios.
In this setting, the adapter trained on the target task is unavailable, but a validation set from the target task is accessible for routing.
Figure \ref{fig:lrvsla} presents the results, with detailed numerical values reported in \Cref{tab:original_adapter_results_with_full_fusion} (Appendix~\ref{app:per_task_group_results}).
With access to validation data, \sys can more accurately identify relevant adapters using task representations.
For \loraretriever, we continue to report OOD performance, as it does not leverage validation data and relies solely on training data.
We find that \sys improves to 92.7\% in the semi-OOD setting from 88.4\% in the OOD setting on Llama2-7B.
Similarly, it improves to 91.1\% from 86.8\% on Llama2-13B.
On both models, this corresponds to gains exceeding 4 points and remains substantially higher than \loraretriever.

\paragraph{Scaling the adapter pool.}
We study the robustness of \sys under large and heterogeneous adapter pool fetched ``\textit{from the wild}''.
We collect all adapters available for Llama2-7B with rank $\leq 64$ from \textsc{HuggingFace}, resulting in a pool of 1567 adapters.
We run \sys on this pool which now does \textit{not} contain the 48 well curated adapters but only adapters fetched from the wild.
In the \ac{OOD} setting, \sys impressively achieves an average normalized performance of $85.7\%$ compared to $88.4\%$ obtained using the 48 adapter pool.
Similarly, in the semi-OOD setting, \sys achieves $89.6\%$ compared to $92.7\%$ on the well curated pool.
This result demonstrates the potential of \sys to extract useful routing signals from large, noisy adapter collections with no information about their training data whatsoever.
To the best of our knowledge, this is the first work to evaluate \ac{LoRA} adapter routing using such large collections of publicly available adapters.
In this setting, the baseline methods cannot be meaningfully evaluated, as they either require access to adapter training data or remain computationally expensive to scale to 1500+ adapters. The detailed numerical results are included in \Cref{tab:hf} (Appendix~\ref{app:per_task_group_results}); for completeness, we also report results for a combined adapter pool that augments the \textsc{HuggingFace} adapters with the original 48 curated adapters.

\begin{table}[t!]
    \centering
    \scriptsize %
    \caption{Performance on \ac{OOD} tasks. Both the retrieval and composition components of \sys (LA) individually bring performance improvements over the baseline \loraretriever (LR).}
    \label{tab:ablation}
    \begin{tabular}{l c c c c}
        \toprule
        \multirow{2}{*}{\textbf{Model}} & \textbf{Retrieval} & \textbf{Composition} & \multicolumn{2}{c}{\textbf{Norm. Performance}} \\
        \cmidrule(lr){4-5}
         & & & \textbf{non-OOD} & \textbf{OOD} \\
        \midrule
        \multirow{4}{*}{\textit{Llama2-7B}}
         & LR & LR & 92.9 & 83.2 \\
         & LR & LA & 98.6 & 85.7 \\
         & LA & LR & 96.7 & \textbf{89.9} \\
         & LA & LA & \textbf{101.2} & 88.4 \\
        \midrule
        \multirow{4}{*}{\textit{Llama2-13B}}
         & LR & LR & 95.6 & 85.9 \\
         & LR & LA & \textbf{99.1} & 86.2 \\
         & LA & LR & 96.8 & 86.3 \\
         & LA & LA & 98.8 & \textbf{86.8} \\
        \bottomrule
    \end{tabular}
\end{table}

\begin{table}[t!]
    \centering
    \caption{The effect of composition for both \loraretriever~(LR) and \sys~(LA) on \ac{OOD} tasks.} %
    \label{tab:ablation_composition}
    \scriptsize %
    \begin{tabular}{l c c c}
    \toprule
      \textbf{Model} & \textbf{Method}  & \makecell{\textbf{w/o Composition} \\ \textbf{(K=1)}} & \makecell{\textbf{with Composition} \\ \textbf{(K=3)}} \\
      \midrule
      \multirow{2}{*}{\textit{Llama2-7B}}
        & LR & 78.4 & 83.2 \\
        & LA & 81.8 & 88.4 \\
      \midrule
      \multirow{2}{*}{\textit{Llama2-13B}}
        & LR & 76.7 & 85.9 \\
        & LA & 78.3 & 86.8 \\
        \bottomrule
    \end{tabular}
\end{table}

\subsection{Ablation Studies}
\label{sec:ablation}

We conduct two ablation studies to assess: \emph{(i)} the individual effects of the retrieval and composition components of \sys~(LA), and \emph{(ii)} the benefit of composition compared to selecting only the single best adapter.
For the first ablation, we replace the retrieval and composition components of our baseline \loraretriever~(LR) with the corresponding components of LA.
\Cref{tab:ablation} shows that just using the weighted composition scheme of LA improves LR from 92.9\% to 98.6\% on Llama2-7B (\ac{non-OOD}).
Similarly, using only the retrieval component of LA improves LR from 92.9\% to 96.7\% on the same setup.
The best performance of 101.2\% is achieved when both the components are used.
We observe a similar trend for the \ac{OOD} setup with Llama2-13B.
On the other hand, for LLama2-7B in \ac{OOD} setting and LLama2-13B in \ac{non-OOD} setting, peak performance is achieved when replacing only one component of LR with LA.
Nevertheless, the performance of using both LA components remains very close and emerges as globally best in two out of the four cases.
Corresponding to the second ablation, \Cref{tab:ablation_composition} reports the performance of using the single best adapter $(K=1)$ as retrieved by both LR and LA.
As shown in the table, this settings yields fairly low performance compared to the $K=3$ case, demonstrating the importance of composing multiple relevant adapters for a given query in the \ac{OOD} scenario.

\subsection{\sys with pseudo tasks}
\label{sec:clusters}
We study how \textsc{LoRAuter} behaves when ground-truth task labels are not available for the publicly collected queries for constructing the representative task set.
Instead of relying on the original 48 task labels, we apply K-Means to the validation data representations and partition the data into \(K\) pseudo task clusters.
We repeat the evaluation for different values of \(K\) while keeping the total number of validation samples fixed.
In this setting, the routing mechanism must rely purely on the structure of the data, without access to true task annotations.

\begin{figure}[t]
    \includegraphics{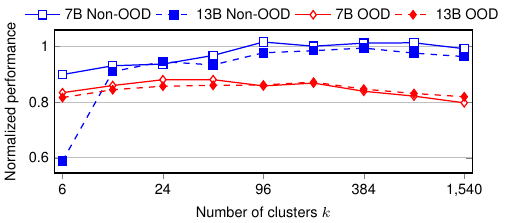}
      \caption{Normalized performance of \textsc{LoRAuter} with varying number of clusters \(K\). %
      }
    \label{fig:cluster_size2}
\end{figure}

Figure~\ref{fig:cluster_size2} shows that the normalized performance remains within a stable range and peaks at an intermediate number of clusters.
Too few clusters produce overly coarse partitions that fail to capture meaningful variation, while too many yield small, noisy clusters that provide unreliable signals for adapter selection.
The optimal choice of \(K\) varies across regimes.
For \ac{OOD} inputs, fewer (larger) clusters perform better, encouraging routing toward robust adapters that generalize across broader data distributions.
In contrast, \ac{non-OOD} inputs benefit from finer-grained clusters that enable selection of more specialized adapters.

Importantly, even without access to ground truth task labels, \sys achieves performance comparable to routing with task labels by appropriately tuning the number of K-Means clusters.
At the best-performing configurations, the method attains up to 101.5\% (7B) and 99.3\% (13B) normalized performance for non-OOD inputs, and 88.0\% (7B) and 87.1\% (13B) for OOD inputs.

\subsection{Searching the best adapter}
\label{sec:adapter_eval}

\begin{figure}[t]
    \includegraphics{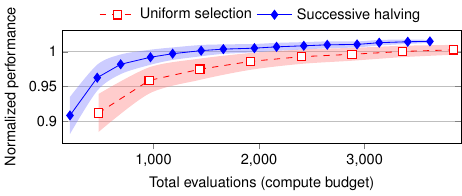}
    \caption{Normalized performance of uniform selection vs.\ Successive Halving  (SH) across evaluation budgets for Llama2-7B.}
    \label{fig:sucessive-halving}
\end{figure}

We now compare two strategies for identifying the best adapter for a given task under a fixed compute budget (the number of adapter calls).
We consider:
\textit{(i) Uniform Selection:} the total evaluation budget is evenly divided across all adapters, and each adapter is evaluated on the same number of validation samples before selecting the highest-scoring one;
\textit{(ii) Successive Halving:} adapters are progressively filtered based on performance estimates, allocating more evaluation budget to promising candidates while respecting the same total budget.
For each budget, we conduct 100 independent runs, where each run randomly samples a subset from the validation datasets.
Figure~\ref{fig:sucessive-halving} shows the normalized performance of both methods across varying evaluation budgets.
Successive Halving (SH) reduces the evaluation budget to reach very close to peak performance by more than 2 times as compared to Uniform Selection.
Thus, \sys scales efficiently to larger adapter pools by avoiding evaluation of all adapters on all samples.
\section{Related work}
\label{sec:rw}

\paragraph{Mixture-of-Experts.}
\ac{MOE}~\citep{jacobs1991adaptive,jordan1994hierarchical,shazeer2017} architectures partition a model into multiple experts and use a learned gating mechanism to sparsely activate experts for each input or token.
This enables expert specialization while scaling model capacity without a proportional increase in inference cost.
\ac{MOE} systems are typically trained end-to-end, with expert parameters and routing functions jointly optimized and fixed after training.
Recent work extends this paradigm to \ac{PEFT} by treating \ac{LoRA} adapters as experts and learning routers to select among them~\citep{muqeeth2024routing,wu2024mixture}.
While effective, these approaches require training additional routing parameters and assume a static set of adapters during router training.
In contrast, \sys is training-free.

\textbf{Model routing.}
Model-routing methods aim to reduce inference cost by dynamically selecting an appropriate model for each input while maintaining response quality.
Prior work~\citep{wang2025mixllm} distinguishes between \emph{non-predictive} and \emph{predictive} routing, where the prior relies on intermediate model outputs to guide routing decisions such as via early exiting~\citep{chen2024frugalgpt,madaan2024automix}, while the latter estimates response quality before inference to directly select a model~\citep{ding2024hybrid,wang2025mixllm}.
While most approaches assume small, static model pools, recent work such as ICL-Router~\citep{wang2025iclrouterincontextlearnedmodel} extends routing to more dynamic settings using learned model capability embeddings.
However, these methods treat models as indivisible units, selecting a single model per query, which is misaligned with modern \ac{PEFT} where a single base model may expose thousands of specialized \ac{LoRA} adapters and where composing multiple adapters can be beneficial.
In contrast, our framework performs predictive routing at the adapter level using task representations and supports input-aware adapter fusion, enabling scalable routing over large adapter collections.

\textbf{Task representations.}
Prior work has explored representing entire tasks, rather than individual inputs, as compact vector representations, commonly referred to as \emph{task representations}.
Such representations have been used to characterize task similarity and predict transferability across tasks~\citep{achille2019task2vec,zamir2018taskonomy}.
More recently, \ac{MOE}-based routing architectures have incorporated task-related signals to guide expert selection~\citep{liang-etal-2025-thor}.
In contrast, existing approaches to \ac{LoRA} adapter routing primarily operate at the level of individual adapters, constructing adapter representations from either their parameters or training data, and do not expose tasks as explicit, reusable routing primitives.
While \loraretriever~\citep{zhao2024loraretriever} implicitly leverages task-related information for input-aware adapter retrieval, tasks are not represented as first-class entities and therefore cannot be independently associated with or shared across adapters.

\section{Conclusions}

We introduced \sys, a training-free framework for scalable \ac{LoRA} adapter routing that operates at the level of tasks rather than individual adapters.
By combining task representations, efficient task–adapter pairing, and input-aware adapter composition, \sys enables mixed-task inference without access to adapter training data and with overhead that scales in the number of tasks.
Empirically, \sys matches oracle task-aligned performance in-domain and achieves state-of-the-art results out-of-domain.
Overall, \sys provides a practical foundation for leveraging large, dynamic adapter pools for \ac{LoRA} routing.

\clearpage

\section*{Impact statement}

This work advances Machine Learning by enabling scalable, training-free routing and composition of large collections of LoRA adapters. By selectively reusing and fusing task-relevant adapters, our framework can improve the performance of smaller base language models, reducing the need to rely on larger and more memory-intensive LLMs. This is particularly beneficial for users operating under GPU memory constraints, where using large models may be impractical or prohibitively expensive. More broadly, our framework is designed to benefit the open-source community by allowing practitioners to reuse and combine thousands of publicly available adapters for new tasks without requiring training data or additional fine-tuning. This can substantially lower computational and financial barriers, making high-quality, task-adaptive LLM capabilities more accessible to a wider range of users, including academic researchers, small labs, and independent developers.

Our method inherits the strengths and weaknesses of the underlying base models and adapters, including potential biases, privacy risks, or harmful behaviors. By making it easier to rapidly reuse and combine public adapters, it could also unintentionally enable misuse in sensitive domains. We therefore encourage responsible deployment, including careful vetting of adapters, appropriate safety filters, and consideration of ethical and regulatory requirements.
\bibliography{references}
\bibliographystyle{unknownconf}

\appendix
\onecolumn

\section{Evaluation details}
\label{appendix:evaluation_details}

This section provides additional evaluation details for all methods considered in our study. All baselines are evaluated under an identical experimental setup: each method operates over the same pool of 48 LoRA adapters with fixed rank \( r = 6 \) and scaling \( \alpha = 12 \), and is assessed on the same mixed-task evaluation set. All experiments are conducted using both \textbf{LLaMA2-7B} and \textbf{LLaMA2-13B} backbone models, with inference performed using \texttt{bfloat16} precision. For \textbf{\sys Fusion}, we fuse the $k = 3$ adapters and use a softmax temperature of \( \tau = 0.2 \) in all evaluations. The sentence transformer used is \url{https://huggingface.co/Styxxxx/lora_retriever}, which is the same as the one found in the \loraretriever paper.

\begin{figure*}[ht]
\centering
\includegraphics[width=0.9\textwidth]{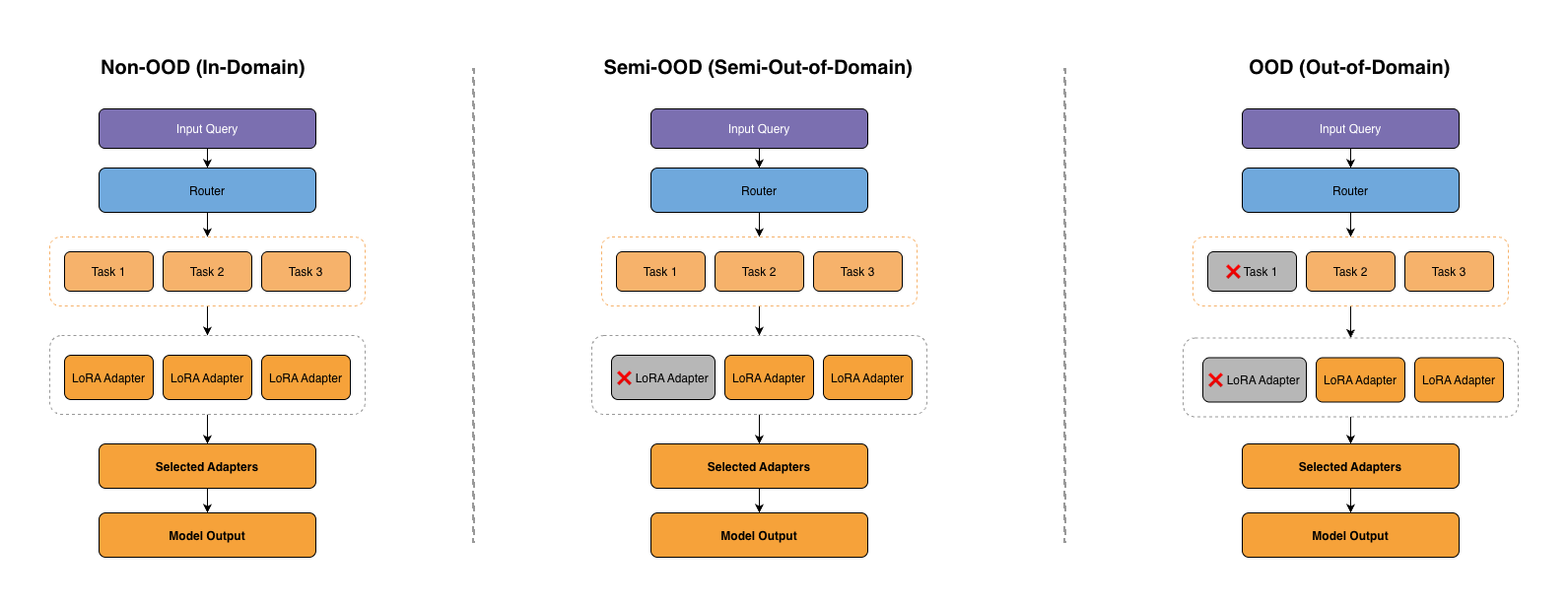}
\caption{
    Illustration of the evaluation regimes used in our experiments.
\ac{non-OOD} (left) corresponds to the in-domain setting where both task-specific validation samples and the task-aligned LoRA adapter are available during routing.
\emph{Semi-OOD} (center) removes the aligned adapter while retaining validation samples.
\ac{OOD} (right) excludes both the task-specific validation samples and the corresponding adapter.
    }
\label{fig:ood_protocol}
\end{figure*}

We evaluate our methods under three different routing regimes, illustrated in Figure~\ref{fig:ood_protocol}. In the \ac{non-OOD} (in-domain) setting, both the validation samples and the LoRA adapter corresponding to the input’s ground-truth task are available during routing. In the \emph{Semi-OOD} setting, validation samples for the target task remain available, but the adapter trained explicitly on that task is removed from the adapter pool, requiring methods to generalize by selecting adapters trained on related tasks. Finally, in the \ac{OOD} (out-of-domain) setting, both the task-specific validation samples and the corresponding adapter are excluded, representing the most challenging scenario in which routing must rely entirely on cross-task generalization.

\subsection{Details of baseline methods}

For completeness, we evaluate \sys against several representative training-free or lightweight adapter–routing approaches. All baselines are implemented following the algorithmic descriptions in their respective papers, and where applicable, we reuse the same text encoder and LoRA configurations to ensure a fair comparison. The baselines correspond exactly to those summarized in Section~4.1 of this paper.

\paragraph{\lorahub} \cite{huang2024lorahub}.
\lorahub is a task-specific adapter–merging method that learns task-specific fusion weights over a set of LoRA parameters via black-box optimization. The method outputs a single fused adapter per task and does not support input-adaptive routing. Following the recommendations in the original paper, we randomly sample 20 adapters from the full pool to constrain the optimization search space. For evaluation, each input is answered using the fused adapter optimized for its corresponding task (i.e., we select the task-specific fused adapter based on the ground-truth task label). LoRAHub was trained using the same validation set as in our method, and we fix the number of optimization iterations to 48, matching the total number of adapter evaluations performed by our approach.

\paragraph{\loraretriever} \cite{zhao2024loraretriever}.
A retrieve-then-compose method that ranks LoRA adapters using similarity between input and adapter representations, then fuses the top-$k$ adapters uniformly. This method assumes the availability of supervised training data for each task and requires homogeneous adapter training conditions. Our method was evaluated using the exact same set of adapters and the same text encoder as employed in the original \loraretriever paper.

\paragraph{\arrow} \cite{ostropenko2024arrow}.
A spectral routing method that applies SVD to adapter representations and selects experts based on spectral affinity with the input. It is training-free but introduces additional per-token inference overhead proportional to the number of adapters. Following the recommendations of the original paper, we use a temperature of $1$ for converting spectral affinities into routing weights, and we retrieve $k = 3$ adapters during routing for consistency with our other baselines.

\paragraph{SpectR} \cite{fleshman2024spectr}.
A recently proposed spectral routing variant that aligns input and adapter representations through eigenspace projections. It avoids training but generally underperforms methods that incorporate task-level context. While the method supports softmax-based weighting of the selected adapters, in our evaluation we follow the configuration used in the original paper and apply uniform weights instead. Furthermore, we retrieve $k=4$ adapters during routing, as suggested in the original paper.

\paragraph{Oracle task-aligned adapter.}
For reference, we report an \emph{oracle task-aligned} baseline, which corresponds to selecting the LoRA adapter that was explicitly trained on the ground-truth task of each input.

\paragraph{Normalized average computation.}
To obtain a single aggregate performance metric across heterogeneous tasks and evaluation scales, we compute a normalized average. For structure-to-text generation, we first average \textsc{Rouge}-1, \textsc{Rouge}-2, and \textsc{Rouge}-L to obtain a single score per method. Then, for each task (i.e., each row in the results table), we normalize all method scores using the corresponding value in the \textit{Oracle Task-Aligned} column as the normalization constant. Finally, the normalized averages reported in Table~2 are computed by averaging these per-task normalized scores across all tasks.

\subsection{Dataset details}
\label{appendix:dataset_details}

We adopt the mixed-task evaluation benchmark introduced by \citet{zhao2024loraretriever}, which is designed to reflect realistic multi-task inference scenarios where user prompts span heterogeneous task types and task labels are not available at inference time. The benchmark comprises a pool of 48 LoRA adapters, each trained independently on a distinct downstream task drawn from a diverse set of natural language understanding (NLU) and natural language generation (NLG) datasets. This setup mirrors practical serving environments in which LoRA adapters are specialized for different tasks.

\begin{figure*}[ht]
\centering
\includegraphics[width=0.9\textwidth]{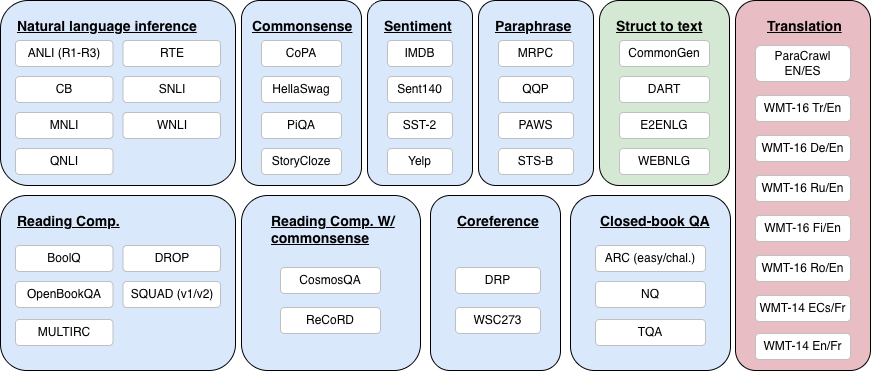}
\caption{
    Overview of the task taxonomy and dataset coverage used in the mixed-task benchmark.
    Each LoRA adapter is trained on a dataset belonging to a specific task cluster (e.g., natural language inference, reading comprehension, translation), following the taxonomy introduced in FLAN~\citep{wei2022finetuned}. At inference time, inputs are provided without task labels, requiring routing methods to infer the underlying task and select or compose suitable adapters.  Figure reproduced from \loraretriever~\citep{zhao2024loraretriever}.
    }
\label{fig:tasktaxonomy}
\vskip -0.2in
\end{figure*}

\paragraph{Task taxonomy.}
The 48 tasks covered by the benchmark follow the high-level categorization popularized by the \textit{FLAN} instruction-tuning framework \citep{wei2022finetuned}. For evaluation and reporting, as illustrated in Figure~\ref{fig:tasktaxonomy} and reported in~\citet{zhao2024loraretriever}, these 48 tasks are grouped into 10 semantically coherent benchmark tasks.
Concretely, the 48 adapters span task categories including \emph{natural language inference}, \emph{reading comprehension}, \emph{closed-book question answering}, \emph{commonsense reasoning}, \emph{sentiment analysis}, \emph{coreference resolution}, \emph{paraphrase detection}, \emph{structure-to-text generation}, and \emph{machine translation}. This taxonomy ensures coverage of both classification-style tasks with discrete label spaces and open-ended generation tasks with free-form outputs.

\paragraph{Instruction formatting.}
Following the instruction-based paradigm of \citet{wei2022finetuned}, each dataset is verbalized using natural language instructions that describe the task objective (e.g., entailment determination, sentiment classification, or translation). While multiple instruction templates are used during adapter training in the original datasets, our evaluation treats adapters as black-box modules and does not assume access to their training prompts or data. At inference time, inputs are provided as plain user queries without explicit task identifiers, reflecting the mixed-task setting.

\paragraph{Validation and test data.}
For each task, a small validation subset is used exclusively for routing or adapter selection, while a disjoint test set is used for final evaluation. Validation sets are intentionally lightweight (typically a few hundred examples per task) to reflect realistic constraints where large-scale labeled data may be unavailable. Test metrics follow standard task-specific evaluation measures (e.g., accuracy, exact match, BLEU, or ROUGE), as reported in the main results.

Overall, this benchmark provides a controlled yet challenging testbed for evaluating scalable LoRA routing methods under both in-domain and distribution-shifted conditions.

\section{Additional results}
\label{appendix:additional_results}

In addition to the primary findings discussed in the main body, this section provides a comprehensive breakdown of the performance of alternative adapter routing strategies. Beyond these comparisons, we also include supplementary adapter evaluation results on the validation set. Finally, we present supplementary results for the Successive Halving (SH) algorithm to further illustrate its efficiency.

\subsection{Selection-based methods}

In this section, we report results for several selection-based strategies for adapter routing. While these methods achieve strong performance in the \ac{non-OOD} setting, they underperform relative to weighted fusion-based routing. This gap arises because selection-based approaches are more sensitive to input–task misalignment introduced by the imperfect text encoder used for input-to-task routing.

Moreover, selection-based methods collapse in the \emph{Semi-OOD} and \ac{OOD} settings, as they lack a mechanism for aggregating knowledge across multiple related tasks and must rely on a single selected adapter. Table~\ref{tab:original_adapter_results_with_selection} summarizes the performance of the evaluated selection strategies across the \ac{non-OOD}, \emph{Semi-OOD}, and \ac{OOD} settings.

The quantitative results in Table~\ref{tab:original_adapter_results_with_selection} highlight a widening performance gap as the evaluation shifts toward out-of-distribution (\ac{OOD}) scenarios. In the \ac{non-OOD} setting, both \loraretriever and \sys selections maintain high normalized averages (e.g., 99.0\% for \sys on Llama2-7B), indicating that the text encoders are generally capable of identifying the correct specialized adapter when the task is known. However, this proficiency diminishes significantly in \emph{Semi-OOD} and \ac{OOD} regimes. For instance, the normalized average for \sys Selection drops from 99.0\% to 81.8\% on Llama2-7B when moving from \ac{non-OOD} to \ac{OOD}. Furthermore, while the larger Llama2-13B model exhibits higher absolute scores across most metrics, it follows a similar degradation pattern to the 7B variant, suggesting that increased parameter scale does not inherently resolve the fundamental limitations of hard-selection routing in zero-shot or \ac{OOD} contexts.

\begin{table*}[t]
\caption{
Performance of selection-based adapter routing methods across evaluation regimes.
\emph{Task-Aligned} reports the oracle adapter trained on the ground-truth task, while \emph{Perf-Aligned} selects the single adapter with highest validation performance on the ground-truth task.
\emph{\loraretriever Selection} and \emph{\sys Selection} denote input-aware single-adapter selection ($k=1$) using the corresponding routing method.
}
\label{tab:original_adapter_results_with_selection}
\begin{center}
\begin{small}
\begin{sc}
\resizebox{0.8\textwidth}{!}{%
\begin{tabular}{
l|c|c|c|c|c|c|c}
\toprule
\multicolumn{8}{c}{\textbf{Llama2-7B}} \\
\midrule
 & \multicolumn{2}{c|} {Oracle}
 & \multicolumn{2}{c|}{Non-OOD}
 & \multicolumn{1}{c|}{Semi-OOD}
 & \multicolumn{2}{c}{OOD}\\
\cmidrule(lr){2-8}
Domain-Metric
& \makecell{Task-\\Aligned}
& \makecell{Perf-\\Aligned}
& \makecell{LoRA-\\Retriever\\Selection}
& \makecell{\sys\\Selection}
& \makecell{\sys\\Selection}
& \makecell{LoRA-\\Retriever\\Selection}
& \makecell{\sys\\Selection} \\
\midrule
struct to text-rouge-1 & 63.5 & 63.5 & 61.3 & 61.0 & 49.1 & 50.1 & 43.4 \\
struct to text-rouge-2 & 39.0 & 39.0 & 37.0 & 36.9 & 25.1 & 26.6 & 20.8 \\
struct to text-rouge-l & 56.6 & 56.6 & 54.5 & 54.6 & 43.2 & 43.9 & 38.5 \\
commonsense-em        & 63.0 & 68.5 & 55.5 & 68.5 & 65.5 & 46.0 & 62.0 \\
sentiment-em          & 90.0 & 90.5 & 89.5 & 90.0 & 87.5 & 89.0 & 90.0 \\
reading comp-em       & 67.7 & 67.7 & 51.7 & 63.7 & 53.7 & 40.3 & 48.7 \\
closed\_book QA-em    & 44.0 & 50.5 & 40.0 & 45.0 & 40.0 & 43.0 & 41.0 \\
coreference-em        & 51.0 & 59.0 & 50.0 & 59.0 & 59.0 & 46.0 & 52.0 \\
read.comp.w:com       & 70.0 & 70.0 & 69.0 & 59.0 & 36.0 & 30.0 & 19.0 \\
paraphrase-em         & 63.5 & 69.0 & 58.0 & 60.5 & 56.0 & 45.5 & 53.0 \\
nli-em                & 73.1 & 72.5 & 70.0 & 71.7 & 67.6 & 60.6 & 64.0 \\
translation-bleu      & 13.1 & 12.6 & 12.8 & 12.6 & 11.9 & 12.0 & 11.7 \\
\midrule
\textbf{Normalized Average (\%)} & 100.0 & 104.4 & 93.2 & 99.0 & 88.4 & 78.4 & 81.8 \\
\midrule
\multicolumn{8}{c}{\textbf{Llama2-13B}} \\
\midrule
 & \multicolumn{2}{c|} {Oracle}
 & \multicolumn{2}{c|}{Non-OOD}
 & \multicolumn{1}{c|}{Semi-OOD}
 & \multicolumn{2}{c}{OOD}\\
\cmidrule(lr){2-8}
Domain-Metric
& \makecell{Task-\\Aligned}
& \makecell{Perf-\\Aligned}
& \makecell{LoRA-\\Retriever\\Selection}
& \makecell{LoRAuter\\Selection}
& \makecell{LoRAuter\\Selection}
& \makecell{LoRA-\\Retriever\\Selection}
& \makecell{LoRAuter\\Selection} \\
\midrule
struct to text-rouge-1 & 65.0 & 65.0 & 62.6 & 62.6 & 50.4 & 49.4 & 44.1 \\
struct to text-rouge-2 & 40.4 & 40.4 & 38.2 & 38.2 & 26.8 & 25.8 & 21.9 \\
struct to text-rouge-l & 58.3 & 58.3 & 56.0 & 56.0 & 44.0 & 42.9 & 38.9 \\
commonsense-em          & 69.0 & 68.5 & 59.0 & 66.0 & 64.0 & 47.5 & 65.5 \\
sentiment-em            & 90.0 & 90.0 & 90.5 & 90.0 & 90.0 & 91.0 & 92.0 \\
reading comp-em         & 76.3 & 76.3 & 60.3 & 72.3 & 60.7 & 48.0 & 46.0 \\
closed\_book QA-em      & 65.0 & 64.0 & 60.0 & 61.0 & 58.0 & 53.0 & 56.5 \\
coreference-em          & 75.0 & 76.0 & 75.0 & 76.0 & 69.0 & 65.0 & 55.0 \\
read.comp.w:com         & 81.0 & 81.0 & 80.0 & 67.0 & 54.0 & 33.0 & 34.0 \\
paraphrase-em           & 76.5 & 76.5 & 68.0 & 67.0 & 59.5 & 52.5 & 53.5 \\
nli-em                  & 82.0 & 82.6 & 78.9 & 79.6 & 69.5 & 70.2 & 73.5 \\
translation-bleu        & 12.8 & 13.2 & 12.9 & 13.2 & 12.9 & 12.7 & 12.8 \\
\midrule
\textbf{Normalized Average (\%)} & 100.0 & 100.3 & 93.8 & 95.2 & 85.8 & 76.7 & 78.3 \\
\bottomrule
\end{tabular}
}
\end{sc}
\end{small}
\end{center}
\vskip -0.2in
\end{table*}

\paragraph{Varying the number of task clusters.}
In this section, we study how sensitive \sys selection is to the granularity of the task database by varying the number of pseudo-task clusters \(K\) used during routing. Following Section~4.4, we remove access to ground-truth task labels and instead apply \textit{K-Means} on validation-set representations, partitioning the available validation data into \(K\) clusters while keeping the total validation budget fixed. \sys then performs retrieval and adapter pairing with respect to these cluster-defined pseudo-tasks. As shown in Figure~6, performance exhibits an inverted-U trend across model sizes and evaluation regimes: using too few clusters produces overly coarse groupings that blur meaningful task structure, whereas using too many clusters yields small, noisy partitions that provide unreliable signals for adapter selection. Moreover, the best-performing \(K\) shifts slightly by regime, with the \ac{OOD} setting favoring fewer (larger) clusters than \ac{non-OOD}. Under distribution shift \sys benefits from selecting \emph{robust} adapters that perform well across a \emph{broad} set of validation samples. In contrast, in the \ac{non-OOD} regime finer-grained clustering enables \sys to pick more \emph{specialized} adapters that perform well on a small set of samples.

Under our cluster-based evaluation, the best \(K\) yields normalized scores of \(98.7\%\) (\(K{=}96\)), and \(97.4\%\) (\(K{=}384\)) for LLaMA2-7B and LLaMA2-13B in the \ac{non-OOD} setting, while dropping to \(85.1\%\), \((K{=}12\)) and \(81.7\%\), (\(K{=}12\)) in \ac{OOD} setting. These trends align with the results in \ref{tab:original_adapter_results_with_selection}  for \textbf{\sys Selection}, which report \(99.0\%\) (7B) and \(95.2\%\) (13B) in \ac{non-OOD} versus \(81.8\%\) (7B) and \(78.3\%\) (13B) in \ac{OOD}.

\begin{figure}[ht]
\centering

    \includegraphics{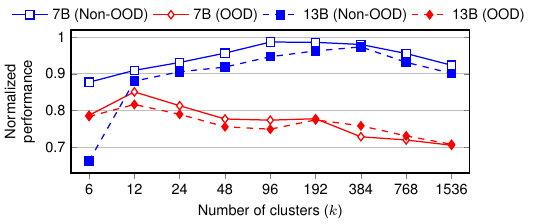}
    \caption{Normalized performance of \sys selection method ($k=1$) as the number of clusters \(K\) varies for both 7B and 13B models. \sys almost reaches the upper bound set by the oracle.
    }
    \label{fig:cluster_size}
\end{figure}

\subsection{\loraretriever weighted fusion}

This subsection evaluates the effect of applying our weighted fusion mechanism to the \loraretriever routing strategy. Table~\ref{tab:original_adapter_results_with_weighted_mixture} reports results across \ac{non-OOD} and \ac{OOD} regimes for both LLaMA2-7B and LLaMA2-13B backbones. Incorporating weighted fusion substantially improves the previous \loraretriever Mixture method, and achieves performance that is nearly on par with \textbf{\sys} in both the \ac{non-OOD} and \ac{OOD} settings. However, unlike \sys, this approach does not exploit additional task samples when they are available. As a result, in the \emph{Semi-OOD} setting, where limited validation data from the target task can be leveraged for routing, weighted fusion over \loraretriever cannot adapt its mixture to the new task and therefore significantly underperforms the \sys method.

Comparing normalized averages across the three fusion strategies further clarifies this trend. In \ac{non-OOD}, \loraretriever's uniform mixture reaches \(92.9\%\) (LLaMA2-7B) and \(95.6\%\) (LLaMA2-13B), while weighted fusion improves to \(98.6\%\) and \(99.1\%\), approaching \sys Fusion at \(101.2\%\) and \(98.8\%\), respectively. In \ac{OOD}, the same ordering holds: the original mixture scores \(83.2\%\) (7B) and \(85.9\%\) (13B), weighted fusion rises to \(85.7\%\) and \(86.2\%\), and \sys Fusion remains strongest at \(88.4\%\) and \(86.8\%\). Overall, weighting reduces errors from imperfect retrieval by upweighting the most relevant retrieved adapters, yielding consistent gains over uniform mixing.

\begin{table*}[h]
\caption{Results of applying weighted adapter fusion to \loraretriever on Non-OOD and OOD tasks for LLaMA2-7B and LLaMA2-13B}
\label{tab:original_adapter_results_with_weighted_mixture}
\begin{center}
\begin{small}
\begin{sc}
\resizebox{0.8\textwidth}{!}{%
\begin{tabular}{
l|
C{2.2cm}|C{2.2cm}|
C{2.2cm}|C{2.2cm}
}
\toprule
 & \multicolumn{2}{c|}{\textbf{Llama2-7B}}
 & \multicolumn{2}{c}{\textbf{Llama2-13B}} \\
\cmidrule(lr){2-3} \cmidrule(lr){4-5}
Domain-Metric
& \makecell{Non-OOD}
& \makecell{OOD}
& \makecell{Non-OOD}
& \makecell{OOD} \\
\midrule
struct to text-rouge-1 & 59.5 & 48.0 & 60.7 & 50.3 \\
struct to text-rouge-2 & 33.6 & 24.7 & 36.2 & 27.1 \\
struct to text-rouge-l & 52.7 & 42.5 & 53.6 & 43.9 \\
commonsense-em        & 60.0 & 58.0 & 70.0 & 67.5 \\
sentiment-em          & 90.5 & 91.0 & 92.0 & 92.5 \\
reading comp-em       & 59.3 & 47.7 & 70.3 & 53.0 \\
closed\_book QA-em    & 47.0 & 46.5 & 66.0 & 59.0 \\
coreference-em        & 58.0 & 57.0 & 81.0 & 62.0 \\
read.comp.w:com       & 66.0 & 36.0 & 79.0 & 56.0 \\
paraphrase-em         & 62.5 & 45.5 & 71.5 & 58.0 \\
nli-em                & 71.0 & 63.9 & 80.0 & 74.9 \\
translation-bleu      & 13.2 & 12.2 & 13.4 & 13.9 \\
\midrule
\textbf{Normalized Average (\%)} & 98.6 & 85.7 & 99.1 & 86.2 \\
\bottomrule
\end{tabular}
}
\end{sc}
\end{small}
\end{center}
\vskip -0.1in
\end{table*}

\subsection{Main results per benchmark task}
\label{app:per_task_group_results}

\paragraph{OOD, non-OOD, semi-OOD settings.}
We complement Figure~\ref{fig:normalized_acc} by reporting results broken down by benchmark task. Table~\ref{tab:original_adapter_results_with_full_fusion} and Table~\ref{tab:baselines_perfomance} present per-benchmark normalized scores, showing that \sys achieves the strongest performance on the majority of tasks. This indicates that the gains observed in Figure~\ref{fig:normalized_acc} are not driven by a single task category, but are broadly consistent across diverse benchmarks.

We further observe a similar trend in the semi-OOD setting, where the task-specific adapter is unavailable but validation data for the target task remains accessible. In this regime, \sys continues to outperform baselines, demonstrating its ability to leverage task-level validation signals. Compared to the fully OOD setting, semi-OOD performance improves consistently across benchmarks. Overall, these results confirm that \sys’s advantages hold at a fine-grained, per-task level and extend robustly across non-OOD, semi-OOD, and OOD scenarios.

\paragraph{Scaling the adapter pool.}
The scaling results in \Cref{sec:main_results} show that \sys maintains performance comparable to baseline routing methods evaluated on the 48-adapter pool, even when operating with adapters collected exclusively ``from the wild,'' over substantially larger and noisier adapter collections, and without access to well-curated task-specific adapters.

In addition to the HF-only setting discussed in \Cref{sec:main_results}, we also evaluate \sys on a combined adapter pool consisting of the 1{,}567 publicly available \textsc{HuggingFace} adapters augmented with the original 48 curated task-specific adapters.
This setting allows us to assess whether \sys can effectively leverage high-quality task-aligned adapters when they are present, while remaining robust to a large number of adapters.
We find that performance in this combined pool closely matches the results obtained with the curated adapters alone, indicating that the presence of a large, noisy adapter pool does not hinder \sys’s ability to select and compose strong adapters.
Detailed per-task results are provided in \Cref{tab:hf} and \Cref{tab:results_hf_average}.

\FloatBarrier
\begin{table*}[t]
\caption{Adapter performance across multiple domains.
\textit{Non-OOD} denotes in-domain tasks, while \textit{OOD} and \textit{Semi-OOD} correspond to different out-of-domain evaluation settings.
The \sys column represents the proposed input-aware adapter routing method. The \textit{Oracle Task-Aligned} baseline (shaded in gray) reports performance of the adapter trained on the ground-truth task of each input.
Results for the \textit{\loraretriever Mixture} are taken directly from the original \loraretriever paper.
All average scores are normalized by the Task-Aligned performance per task.}
\label{tab:original_adapter_results_with_full_fusion}

\begin{center}
\begin{small}
\begin{sc}
\resizebox{0.8\textwidth}{!}{%
\begin{tabular}{
l|c|c|
>{\columncolor{lorautercol}}c|
c|
>{\columncolor{lorautercol}}c|
>{\columncolor{lorautercol}}c}
\toprule
\multicolumn{7}{c}{\textbf{Llama2-7B}} \\
\midrule
 & \multicolumn{1}{c|}{}
 & \multicolumn{2}{c|}{Non-OOD}
 & \multicolumn{2}{c|}{OOD}
 & \multicolumn{1}{c}{Semi-OOD} \\
\cmidrule(lr){3-7}
Domain-Metric
& \makecell{Oracle\\Task-\\Aligned}
& \makecell{LoRA-\\Retriever\\Mixture}
& \makecell{LoRAuter}
& \makecell{LoRA-\\Retriever\\Mixture}
& \makecell{LoRAuter}
& \makecell{LoRAuter} \\
\midrule
struct to text-rouge-1 & \cellcolor[gray]{0.9}63.5 & 55.9 & \textbf{58.8} & \textbf{50.4} & 48.5 & 51.4\\
struct to text-rouge-2 & \cellcolor[gray]{0.9}39.0 & 30.0 & \textbf{33.6} & \textbf{26.9} & 25.3 & 27.3 \\
struct to text-rouge-l & \cellcolor[gray]{0.9}56.6 & 49.5 & \textbf{51.8} & \textbf{44.0} & 43.1 & 45.3 \\
commonsense-em         & \cellcolor[gray]{0.9}63.0 & 61.5 & \textbf{68.0} & 50.0 & \textbf{63.0} & 65.5 \\
sentiment-em           & \cellcolor[gray]{0.9}90.0 & 89.5 & \textbf{90.5} & 90.5 & \textbf{91.0} & 91.0 \\
reading comp-em        & \cellcolor[gray]{0.9}67.7 & 51.3 & \textbf{62.3} & \textbf{47.3} & \textbf{47.3} & 51.7 \\
closed\_book QA-em     & \cellcolor[gray]{0.9}44.0 & 45.0 & \textbf{50.5} & \textbf{48.5} & 47.5 & 46.0 \\
coreference-em         & \cellcolor[gray]{0.9}51.0 & 63.0 & \textbf{63.0} & 49.0 & \textbf{54.0} & 65.0 \\
read.comp.w:com        & \cellcolor[gray]{0.9}70.0 & 46.0 & \textbf{67.0} & \textbf{40.0} & 39.0 & 42.0 \\
paraphrase-em          & \cellcolor[gray]{0.9}63.5 & 56.5 & \textbf{60.0} & 46.0 & \textbf{53.5} & 57.5 \\
nli-em                 & \cellcolor[gray]{0.9}73.1 & 67.9 & \textbf{70.3} & 56.5 & \textbf{64.0} & 69.3 \\
translation-bleu       & \cellcolor[gray]{0.9}13.1 & \textbf{12.8} & 12.6 & 12.2 & \textbf{12.9} & 11.8 \\
\midrule
\textbf{Normalized Average (\%)} & \cellcolor[gray]{0.9}100.0 & 92.9 & \textbf{101.2} & 83.2 & \textbf{88.4} & 92.7 \\
\midrule
\multicolumn{7}{c}{\textbf{Llama2-13B}} \\
\midrule
 & \multicolumn{1}{c|}{}
 & \multicolumn{2}{c|}{Non-OOD}
 & \multicolumn{2}{c|}{OOD}
 & \multicolumn{1}{c}{Semi-OOD} \\
\cmidrule(lr){3-7}
Domain-Metric
& \makecell{Oracle\\Task-\\Aligned}
& \makecell{LoRA-\\Retriever\\Mixture}
& \makecell{LoRAuter}
& \makecell{LoRA-\\Retriever\\Mixture}
& \makecell{LoRAuter}
& \makecell{LoRAuter} \\
\midrule
struct to text-rouge-1 & \cellcolor[gray]{0.9}65.0 & 57.7 & \textbf{61.1} & \textbf{52.1} & 50.7 & 53.3 \\
struct to text-rouge-2 & \cellcolor[gray]{0.9}40.4 & 32.6 & \textbf{36.5} & \textbf{28.1} & 27.2 & 29.3 \\
struct to text-rouge-l & \cellcolor[gray]{0.9}58.3 & 50.8 & \textbf{53.9} & \textbf{45.4} & 43.8 & 47.0 \\
commonsense-em         & \cellcolor[gray]{0.9}69.0 & 64.0 & \textbf{70.5} & 60.5 & \textbf{67.0} & 69.5 \\
sentiment-em           & \cellcolor[gray]{0.9}90.0 & \textbf{91.5} & \textbf{91.5} & 91.5 & \textbf{92.5} & 91.0 \\
reading comp-em        & \cellcolor[gray]{0.9}76.3 & 60.3 & \textbf{71.3} & 51.3 & \textbf{52.3} & 60.7 \\
closed\_book QA-em     & \cellcolor[gray]{0.9}65.0 & \textbf{63.0} & \textbf{63.0} & \textbf{61.0} & 59.5 & 60.0 \\
coreference-em         & \cellcolor[gray]{0.9}75.0 & 76.0 & \textbf{77.0} & 64.0 & \textbf{66.0} & 68.0 \\
read.comp.w:commonsense-em & \cellcolor[gray]{0.9}81.0 & 78.0 & \textbf{79.0} & \textbf{58.0} & 56.0 & 63.0 \\
paraphrase-em          & \cellcolor[gray]{0.9}76.5 & 71.0 & \textbf{71.0} & 55.5 & \textbf{59.0} & 66.0 \\
nli-em                 & \cellcolor[gray]{0.9}82.0 & 78.1 & \textbf{79.5} & \textbf{75.7} & 74.7 & 75.7 \\
translation-bleu       & \cellcolor[gray]{0.9}12.8 & \textbf{14.6} & 14.3 & \textbf{14.1} & 13.9 & 13.9 \\
\midrule
\textbf{Normalized Average (\%)} & \cellcolor[gray]{0.9}100.0 & 95.6 & \textbf{98.8} & 85.9 & \textbf{86.8} & 91.0 \\
\bottomrule
\end{tabular}
}
\end{sc}
\end{small}
\end{center}
\end{table*}

\begin{table*}[t]
\caption{Per-domain results for \sys on Llama2-7B across Non-OOD, Semi-OOD and OOD settings for two different adapter pools: only 48 well curated adapters (48), only adapters from the wild (HF), both well curated and from the wild adapters (HF+48)}
\label{tab:hf}
\begin{center}
\begin{small}
\begin{sc}
\resizebox{0.7\textwidth}{!}{%
\begin{tabular}{l|c c c|c c }
\toprule
& \multicolumn{3}{c|}{HF+48} & \multicolumn{2}{c}{HF} \\
\cmidrule(lr){2-4}\cmidrule(lr){5-6}
Domain-Metric & Non-OOD & Semi-OOD & OOD & Semi-OOD & OOD \\
\midrule
struct to text-rouge-1   & 58.8 & 51.2 & 49.6 & 48.7 & 36.2 \\
struct to text-rouge-2   & 33.6 & 27.1 & 25.6 & 24.2 & 18.1 \\
struct to text-rouge-l   & 52.0 & 44.7 & 43.9 & 43.5 & 32.0 \\
commonsense-em           & 68.0 & 65.5 & 64.0 & 65.5 & 63.5 \\
sentiment-em             & 90.5 & 90.5 & 91.5 & 89.0 & 88.5 \\
reading comp-em          & 61.7 & 53.3 & 48.3 & 51.0 & 49.7 \\
closed\_book QA-em       & 48.5 & 46.0 & 47.0 & 41.5 & 37.0 \\
coreference-em           & 60.0 & 59.0 & 54.0 & 62.0 & 64.0 \\
read.comp.w:com          & 68.0 & 44.0 & 46.0 & 34.0 & 37.0 \\
paraphrase-em            & 61.0 & 57.0 & 52.5 & 38.5 & 52.5 \\
nli-em                   & 69.9 & 68.2 & 68.0 & 69.6 & 70.3 \\
translation-bleu         & 12.4 & 10.9 & 11.8 & 11.5 & 5.2 \\
\bottomrule
\end{tabular}
}
\end{sc}
\end{small}
\end{center}
\end{table*}

\FloatBarrier
\begin{table}[t]
\caption{Normalized average performance results for \sys on Llama2-7B across Non-OOD, Semi-OOD and OOD settings for three adapter pools: only 48 well curated adapters (48), only adapters from the wild (HF), both well curated and from the wild adapters (HF+48)}
\centering
\label{tab:results_hf_average}
\small
\begin{tabular}{lccc}
\toprule
\textbf{Setting} & \textbf{Non-OOD} & \textbf{Semi-OOD} & \textbf{OOD} \\
\midrule
48        & 101.2 & 92.7 & 88.4 \\
HF        & /   & 89.6 & 85.7 \\
HF + 48   & 100.2 & 91.4 & 89.6 \\
\bottomrule
\end{tabular}
\end{table}

\begin{table*}[t]
\caption{Baseline adapter performance across multiple domains.
\textit{Non-OOD} denotes in-domain tasks, while \textit{OOD} correspond to out-of-domain evaluation setting.}
\label{tab:baselines_perfomance}
\begin{center}
\begin{small}
\begin{sc}
\resizebox{0.8\textwidth}{!}{%
\begin{tabular}{l|c c c|c c c}
\toprule
\multicolumn{7}{c}{\textbf{Llama2-7B}} \\
\midrule
& \multicolumn{3}{c|}{Non-OOD} & \multicolumn{3}{c}{OOD} \\
\cmidrule(lr){2-4}\cmidrule(lr){5-7}
Domain-Metric & LoRAHub & ARROW & SpectR & LoRAHub & ARROW & SpectR \\
\midrule
struct to text-rouge-1   & 55.0 & 39.5 & 44.2 & 44.8 & 40.1 & 44.2 \\
struct to text-rouge-2   & 31.9 & 19.6 & 22.5 & 23.1 & 20.1 & 22.5 \\
struct to text-rouge-l   & 48.0 & 35.5 & 39.5 & 39.8 & 35.6 & 39.5 \\
commonsense-em           & 44.5 & 59.0 & 47.0 & 42.5 & 54.0 & 47.0 \\
sentiment-em             & 61.0 & 82.5 & 69.5 & 63.5 & 86.0 & 69.5 \\
reading comp-em          & 44.0 & 51.7 & 43.0 & 45.0 & 50.7 & 43.0 \\
closed\_book QA-em       & 34.5 & 46.5 & 37.0 & 34.0 & 47.0 & 37.0 \\
coreference-em           & 43.0 & 60.0 & 57.0 & 46.0 & 59.0 & 57.0 \\
read.comp.w:com          & 15.0 & 31.0 & 25.0 & 14.0 & 27.0 & 25.0 \\
paraphrase-em            & 35.0 & 45.0 & 42.5 & 38.0 & 49.5 & 42.5 \\
nli-em                   & 49.2 & 56.2 & 53.5 & 50.6 & 59.9 & 53.5 \\
translation-bleu         & 11.9 & 11.6 & 11.6 & 11.7 & 10.8 & 11.6 \\
\midrule
\textbf{Normalized Average (\%)} & 68.6 & 82.5 & 74.2 & 67.8 & 82.0 & 66.3 \\
\midrule
\multicolumn{7}{c}{\textbf{Llama2-13B}} \\
\midrule
& \multicolumn{3}{c|}{Non-OOD} & \multicolumn{3}{c}{OOD} \\
\cmidrule(lr){2-4}\cmidrule(lr){5-7}
Domain-Metric & LoRAHub & ARROW & SpectR & LoRAHub & ARROW & SpectR \\
\midrule
struct to text-rouge-1   & 57.1 & 45.3 & 48.0 & 44.4 & 46.1 & 48.0 \\
struct to text-rouge-2   & 34.8 & 23.6 & 25.0 & 22.9 & 23.8 & 25.0 \\
struct to text-rouge-l   & 50.9 & 40.1 & 42.0 & 39.0 & 39.9 & 42.0 \\
commonsense-em           & 55.5 & 68.5 & 59.5 & 56.0 & 64.0 & 59.5 \\
sentiment-em             & 43.0 & 90.5 & 75.0 & 41.5 & 91.0 & 75.0 \\
reading comp-em          & 47.3 & 55.7 & 51.7 & 48.0 & 53.0 & 51.7 \\
closed\_book QA-em       & 59.0 & 64.5 & 54.5 & 59.5 & 62.5 & 54.5 \\
coreference-em           & 45.0 & 61.0 & 53.0 & 48.0 & 61.0 & 53.0 \\
read.comp.w:com          & 30.0 & 59.0 & 39.0 & 30.0 & 52.0 & 39.0 \\
paraphrase-em            & 38.5 & 52.5 & 44.5 & 36.5 & 54.5 & 44.5 \\
nli-em                   & 48.5 & 70.2 & 59.8 & 49.6 & 70.7 & 59.8 \\
translation-bleu         & 13.7 & 14.1 & 14.0 & 13.5 & 13.1 & 14.0 \\
\midrule
\textbf{Normalized Average (\%)} & 68.2 & 85.7 & 75.1 & 66.1 & 83.2 & 69.9 \\
\bottomrule
\end{tabular}
}
\end{sc}
\end{small}
\end{center}
\end{table*}
\FloatBarrier

\subsection{Adapter performance on the validation set}

\begin{figure}[ht]
  \centering
  \includegraphics{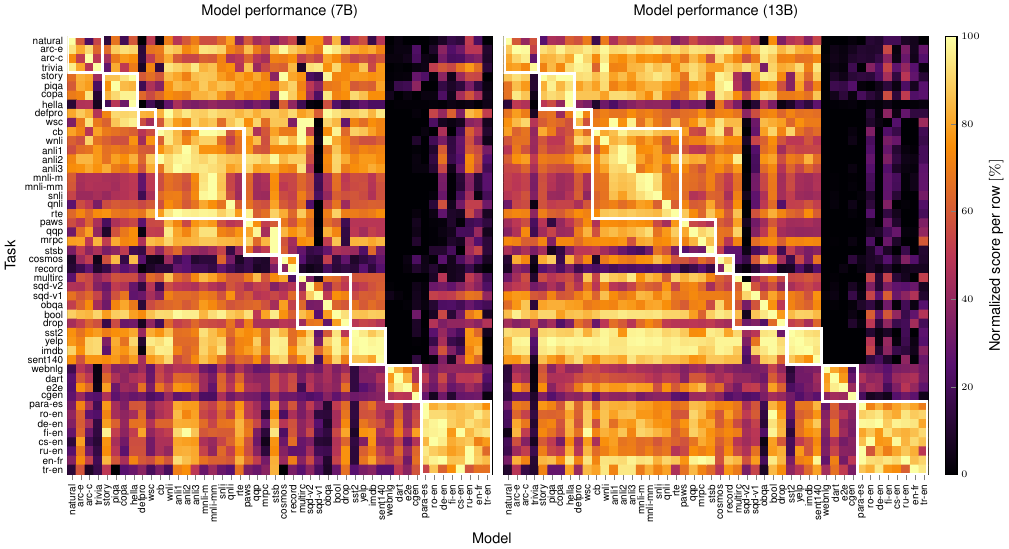}
  \caption{Performance matrix of adapters across tasks evaluated on the validation data. Each row corresponds to one task, while each column represents a single LoRA adapter trained for a specific task.}
  \label{fig:validation_set_perf}
\end{figure}

In this subsection, we analyze the normalized performance of each adapter evaluated on the validation set. Figure~\ref{fig:validation_set_perf} visualizes this performance matrix for both \textbf{LLaMA2-7B} and \textbf{LLaMA2-13B} backbone models. Each row corresponds to a single evaluation task, while each column represents a single LoRA adapter trained for a specific task.

To facilitate comparison across tasks with heterogeneous metrics and score ranges, we apply row-wise normalization. Specifically, for each task (row), we subtract the minimum observed score across all adapters and divide by the range between the maximum and minimum scores for that task, resulting in values normalized to the interval $[0, 1]$. This normalization emphasizes relative adapter performance within each task while removing scale effects induced by differing evaluation metrics.

The figure reveals a pronounced diagonal trend, indicating that adapters trained on a given task generally achieve strong performance when evaluated on that same task. However, this diagonal dominance is not absolute: in several cases, adapters trained on related tasks outperform the task-specific adapter.

In addition to task-specific peaks, we observe consistent blocks of elevated performance among adapters belonging to the same task group. This effect is especially visible for the \emph{structure-to-text} and \emph{machine translation} tasks in the bottom-right of the figure, where adapters form well-defined blocks with strong intra-group generalization, reflecting both their shared structure and their use of different evaluation metrics compared to classification-style tasks.

\subsection{Successive halving}

To efficiently evaluate the effectiveness of \emph{Successive Halving} (SH) for adapter selection, we reuse pre-computed adapter outputs obtained from up to 200 validation samples per task. We exclude tasks with fewer than 200 validation examples to ensure that every run can draw sufficiently large subsets while keeping the evaluation protocol consistent across tasks. To compare performance across heterogeneous metrics and scales, we normalize each task by the same Oracle Task-Aligned domain normalization constants as before, and report the resulting normalized average across tasks.

We estimate the SH and uniform evaluation curves using 100 independent runs per evaluation budget. In each run, we randomly sample subsets of varying sizes from the original pool of 200 validation samples, perform adapter selection under the corresponding budget, and then evaluate the selected adapter on the task’s held-out test set.

\begin{figure}[ht]
    \centering
    \begin{minipage}{0.47\textwidth}
        \centering
        \includegraphics{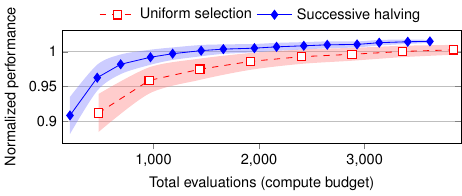}
    \end{minipage}
    \hfill
    \begin{minipage}{0.47\textwidth}
        \centering
        \includegraphics{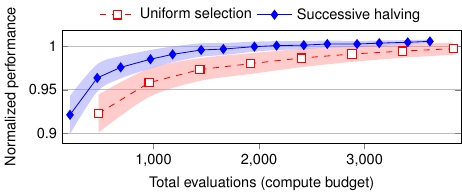}
    \end{minipage}
    \caption{Normalized performance of Successive Halving over uniform selection with respect to their allocated compute budget, for the LLama-7B model (left) and 13B model (right). Standard deviation is reported over the randomness of the validation set ordering with 100 samples.}

    \label{fig:halving-vs-naive}
    \vskip -0.2in
\end{figure}

Figure~\ref{fig:halving-vs-naive} shows that for both \textbf{LLaMA2-7B} and \textbf{LLaMA2-13B}, Successive Halving consistently achieves comparable normalized test performance to uniform evaluation while requiring substantially fewer validation evaluations. In particular, SH reduces the average number of evaluations per adapter by more than \(\times 2\), demonstrating that tournament-style elimination can significantly lower selection cost without sacrificing downstream performance.

\paragraph{Pseudo-code.}
Algorithm~\ref{alg:successive-halving} summarizes the Successive Halving (SH) procedure used for adapter selection under a fixed evaluation budget (measured in adapter-sample evaluations). Compared to the standard formulation, we (i) use a fixed per-adapter evaluation budget for the first $k$ rounds (warmup), and (ii) decouple the elimination rate from the budget schedule by using separate constants for retention ($\eta$) and per-round budget growth ($\gamma$). Starting from an initial candidate set, SH evaluates each adapter on a validation subset, retains the top fraction, and repeats with an increased budget allocation until a single adapter remains.

\begin{algorithm}[htb]
   \caption{Successive Halving}
   \label{alg:successive-halving}
   \begingroup
   \footnotesize
\begin{algorithmic}
   \STATE {\bfseries Input:} adapters $\mathcal{A}$, validation set $\mathcal{V}$, base samples $m$, keep ratio $\eta\in(0,1)$, budget growth $\gamma>1$, rounds $R$, warmup rounds $k$
   \STATE {\bfseries Output:} selected adapter $a^\star$
   \STATE $\mathcal{S} \leftarrow \mathcal{A}$

   \FOR{$r=0$ {\bfseries to} $R-1$}
      \IF{$r < k$}
         \STATE $m_r \leftarrow m$ \hfill // fixed budget during warmup
      \ELSE
         \STATE $m_r \leftarrow \lceil m \cdot \gamma^{\,r-k+1}\rceil$ \hfill // grow budget after warmup
      \ENDIF
      \STATE $m_r \leftarrow \min(m_r,\ |\mathcal{V}|)$
      \STATE Sample $S_r \subseteq \mathcal{V}$ with $|S_r|=m_r$

      \FORALL{$a \in \mathcal{S}$}
         \STATE $s[a] \leftarrow \mathrm{Score}(a, S_r)$
      \ENDFOR

      \STATE $q \leftarrow \max\!\left(1,\left\lceil \eta \cdot |\mathcal{S}| \right\rceil\right)$
      \STATE $\mathcal{S} \leftarrow$ top-$q$ adapters in $\mathcal{S}$ by $s[\cdot]$
      \IF{$|\mathcal{S}| = 1$}
         \STATE {\bfseries break}
      \ENDIF
   \ENDFOR

   \STATE $a^\star \leftarrow$ the single remaining adapter in $\mathcal{S}$
\end{algorithmic}
\endgroup
\end{algorithm}

\subsection{Choosing $K$ for Input-Aware Adapter Fusion}

In Table \ref{tab:k-evaluation}, we evaluate how the number of tasks selected for each input ($K$) affects \sys performance with Llama2-7B in the in-distribution setting. We vary $K$ from $1$ (selecting a single task, i.e., using one adapter without any fusion) to $5$ (fusing the top-$5$ task-specific adapters per input). The best performance is achieved at $K=2$, with $K=3$ performing essentially identically. We therefore select $K=3$ throughout the rest of the paper to remain consistent with prior fusion-based baselines.

\begin{table}[ht]
\centering
\small
\caption{Ablation with different values of $K$. Based on these results, we select $K=3$, which achieves performance close to the best result ($K=2$) while remaining consistent with baselines which also use $K=3$.}

\label{tab:k-evaluation}
\begin{tabular}{lrrrrr}
\toprule
Value of $K$ & 1 & 2 & 3 & 4 & 5 \\
\midrule
struct to text-rouge-1 & \textbf{60.9} & 59.5 & 58.4 & 58.5 & 58.9 \\
struct to text-rouge-2 & \textbf{36.6} & 34.2 & 33.1 & 33.2 & 33.6 \\
struct to text-rouge-l & \textbf{54.4} & 52.9 & 51.3 & 51.8 & 51.7 \\
commonsense-em & 68.0 & \textbf{69.5} & 67.5 & 67.0 & 68.5 \\
sentiment-em & \textbf{91.0} & 90.5 & 90.5 & 90.5 & 90.5 \\
reading comp-em & \textbf{64.0} & 63.7 & 62.7 & 61.0 & 61.3 \\
closed\_book QA-em & 45.0 & 48.5 & 49.5 & 49.0 & \textbf{51.5} \\
coreference-em & 58.0 & 62.0 & \textbf{66.0} & 63.0 & 65.0 \\
read.comp.w:commonsense-em & 59.0 & \textbf{69.0} & 67.0 & 62.0 & 61.0 \\
paraphrase-em & 61.5 & \textbf{62.0} & 60.5 & 60.5 & 61.5 \\
nli-em & \textbf{70.9} & 69.5 & 70.1 & 70.5 & 70.5 \\
translation-bleu & 12.5 & \textbf{12.8} & 12.7 & 12.7 & 12.7 \\
\midrule
Normalized Average (\%) & 98.9 & \textbf{101.8} & 101.6 & 99.9 & 101.2 \\
\bottomrule
\end{tabular}
\end{table}

\pagebreak

\section{Full results}

In Tables \ref{tab:full_results_fusion_7b} and \ref{tab:full_results_fusion_13b}, we report detailed per-task results for \sys across each of the 48 tasks in the evaluation dataset. Results are reported for both LLaMA2-7B and LLaMA2-13B backbone models and cover all evaluation regimes considered in the study.

In Tables \ref{tab:full_results_baselines_7b} and \ref{tab:full_results_baselines_13b}, we report per-task results for the \textsc{LoRAHub}, \textsc{ARROW}, and \textsc{SpectR} baselines on both LLaMA2-7B and LLaMA2-13B, evaluated under the Non-OOD, Semi-OOD, and OOD regimes.

In Table \ref{tab:full_results_hf}, we report per-task results for \sys on Llama2-7B using 1,615 publicly available HuggingFace adapters with rank $\leq 64$, evaluated under Non-OOD, Semi-OOD, and OOD regimes.

In Tables \ref{tab:full_results_selection_7b} and \ref{tab:full_results_selection_13b}, we present detailed per-task results for selection-based routing methods on both LLaMA2-7B and LLaMA2-13B across the same set of 48 evaluation tasks. These tables include results for oracle task-aligned selection, performance-aligned selection, and input-aware selection strategies based on both \loraretriever and \sys.

In Table~\ref{tab:full_results_weighted_mixture}, we present detailed per-task results for the \loraretriever weighted-mixture routing method on both LLaMA2-7B and LLaMA2-13B. All results follow the same evaluation protocol, enabling direct comparison across methods and evaluation settings.

\begin{table*}[t]
\vskip -0.2in
\caption{Per-task results for fusion-based adapter routing methods on LLaMA2-7B across Non-OOD, Semi-OOD, and OOD evaluation regimes. Results for the \loraretriever Mixture baseline are directly copied from the original \loraretriever paper.}
\vskip -0.1in
\label{tab:full_results_fusion_7b}
\begin{center}
\begin{small}
\begin{sc}
\resizebox{0.8\textwidth}{!}{%
\begin{tabular}{l|c|cc|c|cc}
\toprule
 & \multicolumn{1}{c|}{}
 & \multicolumn{2}{c|}{Non-OOD}
 & \multicolumn{1}{c|}{Semi-OOD}
 & \multicolumn{2}{c}{OOD} \\
\cmidrule(lr){3-7}
Domain-Metric
& \makecell{Oracle\\Task-\\Aligned}
& \makecell{LoRA-\\Retriever\\Mixture}
& \makecell{LoRAuter\\Fusion}
& \makecell{LoRAuter\\Fusion}
& \makecell{LoRA-\\Retriever\\Mixture}
& \makecell{LoRAuter\\Fusion} \\
\midrule
\multicolumn{1}{l|}{\textbf{Struct to Text}} & & & & & & \\
web\_nlg\_en-rouge-1 & 69.6 & 57.8 & 63.4 & 52.3 & 53.9 & 49.9 \\
web\_nlg\_en-rouge-2 & 48.0 & 33.5 & 41.9 & 29.0 & 29.4 & 28.0 \\
web\_nlg\_en-rouge-l & 62.8 & 52.3 & 56.3 & 48.1 & 49.6 & 45.4 \\
dart-rouge-1       & 71.7 & 63.2 & 60.1 & 60.3 & 60.0 & 57.4 \\
dart-rouge-2       & 49.4 & 36.6 & 34.4 & 35.7 & 35.4 & 33.6 \\
dart-rouge-l       & 64.9 & 56.3 & 53.1 & 54.1 & 52.4 & 51.8 \\
e2e\_nlg-rouge-1    & 66.1 & 66.0 & 64.1 & 63.1 & 58.7 & 55.6 \\
e2e\_nlg-rouge-2    & 39.6 & 38.8 & 37.6 & 36.0 & 32.1 & 30.4 \\
e2e\_nlg-rouge-l    & 56.4 & 56.9 & 54.0 & 52.9 & 49.0 & 48.6 \\
common\_gen-rouge-1 & 46.5 & 36.5 & 47.8 & 30.1 & 29.0 & 31.0 \\
common\_gen-rouge-2 & 18.9 & 11.1 & 20.6 & 8.4  & 8.6 & 9.2 \\
common\_gen-rouge-l & 42.2 & 32.7 & 43.6 & 26.0 & 24.8 & 26.7 \\
\midrule
\multicolumn{1}{l|}{\textbf{Translation}}  & & & & & & \\
para\_crawl\_enes-bleu        & 24.0 & 22.8 & 23.9 & 19.5 & 22.1 & 23.6 \\
wmt16\_translate\_tren-bleu   & 3.7  & 3.7 & 2.0  & 2.0  & 2.6 & 2.3 \\
wmt16\_translate\_deen-bleu   & 17.8 & 11.0 & 19.5 & 18.5 & 10.8 & 20.4 \\
wmt16\_translate\_ruen-bleu   & 11.5 & 18.8 & 11.9 & 11.3 & 18.7 & 12.0 \\
wmt16\_translate\_fien-bleu   & 6.8  & 7.3 & 7.0  & 5.9  & 7.8 & 6.6 \\
wmt16\_translate\_roen-bleu   & 14.6 & 13.1 & 12.4 & 13.2 & 12.2 & 13.8 \\
wmt14\_enfr-bleu              & 16.0 & 17.8 & 16.8 & 16.8 & 18.0 & 17.9 \\
wmt16\_translate\_csen-bleu   & 10.5 & 8.3 & 7.3  & 7.2  & 5.8 & 6.5 \\
\midrule
\multicolumn{1}{l|}{\textbf{COMMONSENSE}} & & & & & & \\
story\_cloze-em & 72.0 & 84.0 & 96.0 & 96.0 & 58.0 & 88.0 \\
piqa-em        & 44.0 & 38.0 & 44.0 & 38.0 & 34.0 & 42.0 \\
copa-em        & 86.0 & 80.0 & 82.0 & 80.0 & 68.0 & 78.0 \\
hellaswag-em   & 50.0 & 44.0 & 50.0 & 48.0 & 40.0 & 44.0 \\
\midrule
\multicolumn{1}{l|}{\textbf{sentiment}} & & & & & & \\
sst2-em                   & 98.0 & 96.0 & 98.0 & 98.0 & 94.0 & 94.0 \\
yelp\_polarity\_reviews-em & 98.0 & 98.0 & 98.0 & 98.0 & 98.0 & 98.0 \\
imdb\_reviews-em           & 96.0 & 96.0 & 98.0 & 98.0 & 96.0 & 98.0 \\
sentiment140-em           & 68.0 & 68.0 & 68.0 & 70.0 & 74.0 & 74.0 \\
\midrule
\multicolumn{1}{l|}{\textbf{READING Comp.}} & & & & & & \\
multirc-em      & 68.0 & 48.0 & 64.0 & 56.0 & 44.0 & 44.0 \\
squad\_v2-em     & 62.0 & 22.0 & 54.0 & 20.0 & 16.0 & 14.0 \\
squad\_v1-em     & 68.0 & 62.0 & 72.0 & 68.0 & 68.0 & 64.0 \\
openbookqa-em   & 84.0 & 78.0 & 80.0 & 76.0 & 66.0 & 74.0 \\
bool\_q-em       & 84.0 & 80.0 & 82.0 & 74.0 & 76.0 & 74.0 \\
drop-em         & 40.0 & 18.0 & 22.0 & 16.0 & 14.0 & 14.0 \\
\midrule
\multicolumn{1}{l|}{\textbf{CLOSE-BOOK QA}} & & & & & & \\
natural\_questions-em & 18.0 & 16.0 & 14.0 & 12.0 & 10.0 & 12.0 \\
arc\_easy-em          & 48.0 & 66.0 & 78.0 & 76.0 & 82.0 & 82.0 \\
arc\_challenge-em     & 46.0 & 50.0 & 56.0 & 50.0 & 46.0 & 48.0 \\
trivia\_qa-em         & 64.0 & 48.0 & 54.0 & 46.0 & 56.0 & 48.0 \\
\midrule
\multicolumn{1}{l|}{\textbf{COREFERENCE}}  & & & & & & \\
definite\_pronoun\_resolution-em & 52.0 & 68.0 & 70.0 & 72.0 & 56.0 & 58.0 \\
wsc-em                             & 50.0 & 58.0 & 56.0 & 58.0 & 42.0 & 50.0 \\
\midrule
\multicolumn{1}{l|}{\textbf{READ. COMP. W/ COM}}  & & & & & & \\
cosmos\_qa-em & 70.0 & 50.0 & 72.0 & 52.0 & 46.0 & 48.0 \\
record-em    & 70.0 & 42.0 & 62.0 & 32.0 & 34.0 & 30.0 \\
\midrule
\multicolumn{1}{l|}{\textbf{PARAPHRASE}} & & & & & & \\
paws\_wiki-em   & 90.0 & 56.0 & 60.0 & 66.0 & 46.0 & 54.0 \\
glue\_qqp-em    & 70.0 & 80.0 & 78.0 & 78.0 & 58.0 & 74.0 \\
glue\_mrpc-em   & 60.0 & 60.0 & 66.0 & 66.0 & 58.0 & 62.0 \\
stsb-em         & 34.0 & 30.0 & 36.0 & 20.0 & 20.0 & 24.0 \\
\midrule
\multicolumn{1}{l|}{\textbf{NLI}} & & & & & & \\
cb-em               & 88.9 & 86.7 & 88.9 & 86.7 & 66.7 & 57.8 \\
wnli-em             & 72.0 & 60.0 & 56.0 & 58.0 & 54.0 & 50.0 \\
anli\_r1-em          & 52.0 & 40.0 & 54.0 & 50.0 & 42.0 & 48.0 \\
anli\_r2-em          & 46.0 & 46.0 & 46.0 & 44.0 & 46.0 & 48.0 \\
anli\_r3-em          & 46.0 & 44.0 & 46.0 & 46.0 & 50.0 & 48.0 \\
mnli\_matched-em     & 88.0 & 80.0 & 86.0 & 86.0 & 88.0 & 86.0 \\
mnli\_mismatched-em  & 92.0 & 88.0 & 94.0 & 92.0 & 90.0 & 94.0 \\
snli-em              & 96.0 & 90.0 & 84.0 & 84.0 & 92.0 & 82.0 \\
qnli-em              & 94.0 & 74.0 & 70.0 & 68.0 & 38.0 & 66.0 \\
rte-em               & 56.0 & 70.0 & 78.0 & 78.0 & 76.0 & 60.0 \\
\bottomrule
\end{tabular}
}
\end{sc}
\end{small}
\end{center}
\vskip -0.1in
\end{table*}
\begin{table*}[t]
\vskip -0.2in
\caption{Per-task results for fusion-based adapter routing methods on LLaMA2-13B across Non-OOD, Semi-OOD, and OOD evaluation regimes. Results for the \loraretriever Mixture baseline are directly copied from the original \loraretriever paper.}
\vskip -0.1in
\label{tab:full_results_fusion_13b}
\begin{center}
\begin{small}
\begin{sc}
\resizebox{0.8\textwidth}{!}{%
\begin{tabular}{l|c|cc|c|cc}
\toprule
 & \multicolumn{1}{c|}{}
 & \multicolumn{2}{c|}{Non-OOD}
 & \multicolumn{1}{c|}{Semi-OOD}
 & \multicolumn{2}{c}{OOD} \\
\cmidrule(lr){3-7}
Domain-Metric
& \makecell{Oracle\\Task-\\Aligned}
& \makecell{LoRA-\\Retriever\\Mixture}
& \makecell{LoRAuter\\Fusion}
& \makecell{LoRAuter\\Fusion}
& \makecell{LoRA-\\Retriever\\Mixture}
& \makecell{LoRAuter\\Fusion} \\
\midrule
\multicolumn{1}{l|}{\textbf{Struct to Text}} & & & & & & \\
web\_nlg\_en-rouge-1 & 72.4 & 59.7 & 68.9 & 57.4 & 53.7 & 52.4 \\
web\_nlg\_en-rouge-2 & 52.1 & 35.6 & 46.9 & 31.6 & 29.1 & 28.9 \\
web\_nlg\_en-rouge-l & 66.1 & 55.1 & 61.0 & 50.8 & 49.0 & 46.6 \\
dart-rouge-1       & 72.9 & 62.6 & 64.3 & 59.0 & 60.6 & 59.2 \\
dart-rouge-2       & 52.9 & 38.9 & 40.4 & 35.4 & 37.3 & 35.6 \\
dart-rouge-l       & 66.7 & 55.0 & 56.4 & 53.1 & 53.4 & 51.7 \\
e2e\_nlg-rouge-1    & 66.5 & 66.7 & 65.1 & 64.4 & 63.9 & 61.6 \\
e2e\_nlg-rouge-2    & 38.8 & 39.0 & 37.4 & 36.7 & 36.4 & 33.9 \\
e2e\_nlg-rouge-l    & 56.4 & 56.2 & 55.5 & 53.4 & 53.7 & 51.6 \\
common\_gen-rouge-1 & 48.4 & 41.7 & 46.3 & 32.5 & 30.0 & 29.7 \\
common\_gen-rouge-2 & 17.9 & 17.0 & 21.2 & 13.6 & 9.5 & 10.6 \\
common\_gen-rouge-l & 44.1 & 36.7 & 42.8 & 30.5 & 25.3 & 25.4 \\
\midrule
\multicolumn{1}{l|}{\textbf{Translation}}  & & & & & & \\
para\_crawl\_enes-bleu        & 25.0 & 27.4 & 25.9 & 24.3 & 25.8 & 23.6 \\
wmt16\_translate\_tren-bleu   & 2.0  & 3.4 & 3.7  & 3.6  & 2.7 & 2.8 \\
wmt16\_translate\_deen-bleu   & 19.8 & 10.5 & 21.3 & 21.7 & 12.0 & 20.9 \\
wmt16\_translate\_ruen-bleu   & 11.3 & 20.2 & 12.4 & 11.9 & 20.2 & 11.9 \\
wmt16\_translate\_fien-bleu   & 6.8  & 8.7 & 9.3  & 8.2  & 7.8 & 8.6 \\
wmt16\_translate\_roen-bleu   & 14.2 & 17.4 & 11.7 & 11.2 & 13.3 & 13.7 \\
wmt14\_enfr-bleu              & 16.5 & 18.7 & 20.5 & 20.5 & 20.9 & 20.9 \\
wmt16\_translate\_csen-bleu   & 6.9  & 10.3 & 9.5  & 9.7  & 10.1 & 9.0 \\
\midrule
\multicolumn{1}{l|}{\textbf{COMMONSENSE}} & & & & & & \\
story\_cloze-em & 94.0 & 80.0 & 98.0 & 96.0 & 76.0 & 96.0 \\
piqa-em        & 48.0 & 46.0 & 50.0 & 48.0 & 46.0 & 48.0 \\
copa-em        & 76.0 & 78.0 & 80.0 & 82.0 & 76.0 & 80.0 \\
hellaswag-em   & 58.0 & 52.0 & 54.0 & 52.0 & 44.0 & 44.0 \\
\midrule
\multicolumn{1}{l|}{\textbf{sentiment}} & & & & & & \\
sst2-em                   & 98.0 & 98.0 & 98.0 & 96.0 & 100.0 & 100.0 \\
yelp\_polarity\_reviews-em & 98.0 & 98.0 & 98.0 & 98.0 & 98.0 & 100.0 \\
imdb\_reviews-em           & 96.0 & 98.0 & 98.0 & 98.0 & 98.0 & 98.0 \\
sentiment140-em           & 68.0 & 72.0 & 72.0 & 72.0 & 70.0 & 72.0 \\
\midrule
\multicolumn{1}{l|}{\textbf{READING Comp.}} & & & & & & \\
multirc-em      & 88.0 & 66.0 & 78.0 & 66.0 & 44.0 & 40.0 \\
squad\_v2-em     & 68.0 & 60.0 & 68.0 & 38.0 & 60.0 & 18.0 \\
squad\_v1-em     & 74.0 & 34.0 & 70.0 & 70.0 & 24.0 & 64.0 \\
openbookqa-em   & 86.0 & 88.0 & 88.0 & 78.0 & 82.0 & 78.0 \\
bool\_q-em       & 84.0 & 78.0 & 84.0 & 86.0 & 78.0 & 86.0 \\
drop-em         & 58.0 & 36.0 & 40.0 & 26.0 & 20.0 & 28.0 \\
\midrule
\multicolumn{1}{l|}{\textbf{CLOSE-BOOK QA}} & & & & & & \\
natural\_questions-em & 30.0 & 24.0 & 26.0 & 18.0 & 16.0 & 18.0 \\
arc\_easy-em          & 94.0 & 94.0 & 94.0 & 90.0 & 94.0 & 94.0 \\
arc\_challenge-em     & 68.0 & 68.0 & 70.0 & 70.0 & 70.0 & 70.0 \\
trivia\_qa-em         & 68.0 & 66.0 & 62.0 & 62.0 & 64.0 & 56.0 \\
\midrule
\multicolumn{1}{l|}{\textbf{COREFERENCE}}  & & & & & & \\
definite\_pronoun\_resolution-em & 88.0 & 88.0 & 90.0 & 76.0 & 72.0 & 74.0 \\
wsc-em                             & 62.0 & 64.0 & 64.0 & 60.0 & 56.0 & 58.0 \\
\midrule
\multicolumn{1}{l|}{\textbf{READ. COMP. W/ COM}}  & & & & & & \\
cosmos\_qa-em & 84.0 & 82.0 & 82.0 & 80.0 & 70.0 & 60.0 \\
record-em    & 78.0 & 74.0 & 76.0 & 46.0 & 46.0 & 18.0 \\
\midrule
\multicolumn{1}{l|}{\textbf{PARAPHRASE}} & & & & & & \\
paws\_wiki-em   & 92.0 & 88.0 & 88.0 & 78.0 & 74.0 & 64.0 \\
glue\_qqp-em    & 86.0 & 84.0 & 82.0 & 82.0 & 62.0 & 82.0 \\
glue\_mrpc-em   & 84.0 & 78.0 & 72.0 & 68.0 & 64.0 & 64.0 \\
stsb-em        & 44.0 & 34.0 & 42.0 & 36.0 & 22.0 & 26.0 \\
\midrule
\multicolumn{1}{l|}{\textbf{NLI}} & & & & & & \\
cb-em               & 95.6 & 93.3 & 93.3 & 91.1 & 88.9 & 86.7 \\
wnli-em             & 72.0 & 76.0 & 74.0 & 68.0 & 66.0 & 64.0 \\
anli\_r1-em          & 70.0 & 64.0 & 66.0 & 64.0 & 68.0 & 68.0 \\
anli\_r2-em          & 66.0 & 60.0 & 62.0 & 54.0 & 60.0 & 56.0 \\
anli\_r3-em          & 68.0 & 62.0 & 66.0 & 62.0 & 60.0 & 60.0 \\
mnli\_matched-em     & 86.0 & 86.0 & 86.0 & 86.0 & 88.0 & 84.0 \\
mnli\_mismatched-em  & 90.0 & 94.0 & 92.0 & 92.0 & 100.0 & 92.0 \\
snli-em             & 90.0 & 90.0 & 92.0 & 90.0 & 92.0 & 92.0 \\
qnli-em             & 94.0 & 74.0 & 84.0 & 76.0 & 58.0 & 74.0 \\
rte-em              & 88.0 & 82.0 & 80.0 & 74.0 & 76.0 & 70.0 \\
\bottomrule
\end{tabular}
}
\end{sc}
\end{small}
\end{center}
\vskip -0.1in
\end{table*}

\begin{table*}[t]
\caption{Performance of baseline routing methods on Llama2-7B .}
\vskip -0.1in
\label{tab:full_results_baselines_7b}
\begin{center}
\begin{small}
\begin{sc}
\resizebox{0.8\textwidth}{!}{%
\begin{tabular}{l|c c c|c c c}
\toprule
& \multicolumn{3}{c|}{Non-OOD} & \multicolumn{3}{c}{OOD} \\
\cmidrule(lr){2-4}\cmidrule(lr){5-7}
Domain-Metric & LoRAHub & ARROW & SpectR & LoRAHub & ARROW & SpectR \\
\midrule
\multicolumn{1}{l|}{\textbf{Struct to Text}} & & & & & & \\
web\_nlg\_en-rouge-1 & 67.2 & 30.8 & 41.6 & 46.6 & 30.6 & 33.6 \\
web\_nlg\_en-rouge-2 & 44.4 & 15.8 & 22.0 & 24.2 & 15.5 & 16.9 \\
web\_nlg\_en-rouge-l & 59.1 & 29.4 & 39.0 & 42.6 & 28.2 & 31.6 \\
dart-rouge-1         & 61.2 & 49.4 & 53.6 & 60.4 & 50.8 & 44.0 \\
dart-rouge-2         & 36.7 & 26.3 & 30.3 & 36.0 & 29.4 & 24.1 \\
dart-rouge-l         & 54.7 & 44.4 & 48.3 & 53.7 & 46.2 & 39.5 \\
e2e\_nlg-rouge-1      & 65.1 & 49.7 & 53.5 & 47.2 & 52.1 & 49.2 \\
e2e\_nlg-rouge-2      & 38.6 & 27.2 & 29.3 & 25.3 & 27.7 & 26.7 \\
e2e\_nlg-rouge-l      & 55.1 & 42.4 & 45.4 & 40.8 & 43.5 & 42.0 \\
common\_gen-rouge-1   & 26.3 & 28.2 & 28.1 & 24.9 & 26.9 & 25.7 \\
common\_gen-rouge-2   & 7.8 & 9.0 & 8.3 & 6.8 & 7.8 & 6.2 \\
common\_gen-rouge-l   & 23.0 & 25.7 & 25.4 & 22.2 & 24.7 & 24.0 \\
\midrule
\multicolumn{1}{l|}{\textbf{Translation}} & & & & & & \\
para\_crawl\_enes-bleu      & 23.6 & 21.6 & 23.4 & 22.6 & 19.5 & 19.6 \\
wmt16\_translate\_tren-bleu & 3.5 & 2.7 & 2.4 & 2.4 & 3.8 & 2.9 \\
wmt16\_translate\_deen-bleu & 16.9 & 17.6 & 16.6 & 16.4 & 17.5 & 12.6 \\
wmt16\_translate\_ruen-bleu & 10.1 & 9.4 & 9.1 & 9.7 & 6.8 & 6.9 \\
wmt16\_translate\_fien-bleu & 7.3 & 7.2 & 6.6 & 8.4 & 5.9 & 5.9 \\
wmt16\_translate\_roen-bleu & 10.6 & 10.3 & 11.3 & 10.9 & 9.3 & 8.4 \\
wmt14\_enfr-bleu            & 16.6 & 17.1 & 17.0 & 16.9 & 16.8 & 15.0 \\
wmt16\_translate\_csen-bleu & 6.4 & 7.0 & 6.4 & 6.5 & 6.9 & 6.8 \\
\midrule
\multicolumn{1}{l|}{\textbf{COMMONSENSE}} & & & & & & \\
story\_cloze-em & 62.0 & 84.0 & 70.0 & 64.0 & 70.0 & 74.0 \\
piqa-em        & 30.0 & 38.0 & 34.0 & 28.0 & 40.0 & 32.0 \\
copa-em        & 62.0 & 74.0 & 66.0 & 66.0 & 70.0 & 70.0 \\
hellaswag-em   & 24.0 & 40.0 & 18.0 & 12.0 & 36.0 & 12.0 \\
\midrule
\multicolumn{1}{l|}{\textbf{sentiment}} & & & & & & \\
sst2-em                   & 66.0 & 82.0 & 70.0 & 72.0 & 90.0 & 50.0 \\
yelp\_polarity\_reviews-em & 72.0 & 94.0 & 82.0 & 70.0 & 96.0 & 76.0 \\
imdb\_reviews-em           & 50.0 & 92.0 & 70.0 & 52.0 & 96.0 & 62.0 \\
sentiment140-em           & 56.0 & 62.0 & 56.0 & 60.0 & 62.0 & 50.0 \\
\midrule
\multicolumn{1}{l|}{\textbf{READING Comp.}} & & & & & & \\
multirc-em    & 42.0 & 46.0 & 38.0 & 44.0 & 50.0 & 42.0 \\
squad\_v2-em   & 24.0 & 24.0 & 22.0 & 26.0 & 22.0 & 28.0 \\
squad\_v1-em   & 66.0 & 74.0 & 62.0 & 66.0 & 72.0 & 48.0 \\
openbookqa-em & 52.0 & 66.0 & 56.0 & 54.0 & 66.0 & 58.0 \\
bool\_q-em     & 72.0 & 82.0 & 70.0 & 70.0 & 80.0 & 66.0 \\
drop-em       & 8.0 & 18.0 & 10.0 & 10.0 & 14.0 & 4.0 \\
\midrule
\multicolumn{1}{l|}{\textbf{CLOSE-BOOK QA}} & & & & & & \\
natural\_questions-em & 14.0 & 20.0 & 12.0 & 12.0 & 20.0 & 10.0 \\
arc\_easy-em          & 48.0 & 64.0 & 56.0 & 48.0 & 66.0 & 52.0 \\
arc\_challenge-em     & 30.0 & 44.0 & 34.0 & 30.0 & 44.0 & 32.0 \\
trivia\_qa-em         & 46.0 & 58.0 & 46.0 & 46.0 & 58.0 & 36.0 \\
\midrule
\multicolumn{1}{l|}{\textbf{COREFERENCE}} & & & & & & \\
definite\_pronoun\_resolution-em & 50.0 & 68.0 & 62.0 & 56.0 & 66.0 & 58.0 \\
wsc-em                           & 36.0 & 52.0 & 52.0 & 36.0 & 52.0 & 52.0 \\
\midrule
\multicolumn{1}{l|}{\textbf{READ. COMP. W/ COM}} & & & & & & \\
cosmos\_qa-em & 24.0 & 32.0 & 38.0 & 22.0 & 34.0 & 10.0 \\
record-em    & 6.0 & 30.0 & 12.0 & 6.0 & 20.0 & 6.0 \\
\midrule
\multicolumn{1}{l|}{\textbf{PARAPHRASE}} & & & & & & \\
paws\_wiki-em & 46.0 & 48.0 & 42.0 & 44.0 & 60.0 & 48.0 \\
glue\_qqp-em  & 20.0 & 52.0 & 52.0 & 32.0 & 70.0 & 58.0 \\
glue\_mrpc-em & 62.0 & 60.0 & 62.0 & 62.0 & 52.0 & 60.0 \\
stsb-em       & 12.0 & 20.0 & 14.0 & 14.0 & 16.0 & 8.0 \\
\midrule
\multicolumn{1}{l|}{\textbf{NLI}} & & & & & & \\
cb-em               & 44.4 & 75.6 & 71.1 & 57.8 & 73.3 & 80.0 \\
wnli-em             & 44.0 & 52.0 & 54.0 & 46.0 & 48.0 & 46.0 \\
anli\_r1-em          & 36.0 & 32.0 & 36.0 & 34.0 & 38.0 & 34.0 \\
anli\_r2-em          & 34.0 & 32.0 & 36.0 & 34.0 & 40.0 & 28.0 \\
anli\_r3-em          & 44.0 & 38.0 & 40.0 & 44.0 & 40.0 & 36.0 \\
mnli\_matched-em     & 56.0 & 60.0 & 54.0 & 56.0 & 70.0 & 44.0 \\
mnli\_mismatched-em  & 76.0 & 88.0 & 72.0 & 74.0 & 86.0 & 60.0 \\
snli-em              & 64.0 & 70.0 & 62.0 & 64.0 & 74.0 & 58.0 \\
qnli-em              & 46.0 & 58.0 & 52.0 & 46.0 & 68.0 & 56.0 \\
rte-em               & 48.0 & 56.0 & 58.0 & 50.0 & 62.0 & 58.0 \\
\bottomrule
\end{tabular}
}
\end{sc}
\end{small}
\end{center}
\end{table*}

\begin{table*}[t]
\caption{Performance of baseline routing methods on Llama2-13B.}
\vskip -0.1in
\label{tab:full_results_baselines_13b}
\begin{center}
\begin{small}
\begin{sc}
\resizebox{0.8\textwidth}{!}{%
\begin{tabular}{l|c c c|c c c}
\toprule
& \multicolumn{3}{c|}{Non-OOD} & \multicolumn{3}{c}{OOD} \\
\cmidrule(lr){2-4}\cmidrule(lr){5-7}
Domain-Metric & LoRAHub & ARROW & SpectR & LoRAHub & ARROW & SpectR \\
\midrule
\multicolumn{1}{l|}{\textbf{Struct to Text}} & & & & & & \\
web\_nlg\_en-rouge-1 & 71.9 & 39.8 & 47.3 & 42.5 & 37.7 & 41.4 \\
web\_nlg\_en-rouge-2 & 51.9 & 20.1 & 23.5 & 21.9 & 17.6 & 20.1 \\
web\_nlg\_en-rouge-l & 66.1 & 36.3 & 42.3 & 38.9 & 34.0 & 38.2 \\
dart-rouge-1         & 63.0 & 56.5 & 56.5 & 63.0 & 57.3 & 45.6 \\
dart-rouge-2         & 37.8 & 34.1 & 33.6 & 38.0 & 33.2 & 27.2 \\
dart-rouge-l         & 55.9 & 51.5 & 51.1 & 55.7 & 49.6 & 40.2 \\
e2e\_nlg-rouge-1      & 66.9 & 55.9 & 57.6 & 49.3 & 56.6 & 57.1 \\
e2e\_nlg-rouge-2      & 39.1 & 29.5 & 31.0 & 23.8 & 30.6 & 30.3 \\
e2e\_nlg-rouge-l      & 56.6 & 45.4 & 46.5 & 40.0 & 45.2 & 45.7 \\
common\_gen-rouge-1   & 26.7 & 29.0 & 30.4 & 22.9 & 32.8 & 29.3 \\
common\_gen-rouge-2   & 10.3 & 10.6 & 12.0 & 8.1 & 13.7 & 11.4 \\
common\_gen-rouge-l   & 24.9 & 27.4 & 28.2 & 21.4 & 30.9 & 27.3 \\
\midrule
\multicolumn{1}{l|}{\textbf{Translation}} & & & & & & \\
para\_crawl\_enes-bleu      & 22.7 & 24.5 & 26.9 & 22.8 & 23.2 & 23.4 \\
wmt16\_translate\_tren-bleu & 4.0 & 3.9 & 3.3 & 3.3 & 3.6 & 3.3 \\
wmt16\_translate\_deen-bleu & 18.7 & 18.9 & 19.8 & 18.5 & 16.2 & 18.9 \\
wmt16\_translate\_ruen-bleu & 12.2 & 11.9 & 11.2 & 13.0 & 10.0 & 10.5 \\
wmt16\_translate\_fien-bleu & 9.9 & 8.9 & 8.2 & 9.9 & 8.2 & 6.9 \\
wmt16\_translate\_roen-bleu & 12.1 & 13.5 & 12.5 & 11.8 & 13.3 & 10.5 \\
wmt14\_enfr-bleu            & 19.3 & 19.1 & 19.6 & 18.3 & 18.8 & 18.9 \\
wmt16\_translate\_csen-bleu & 11.1 & 11.8 & 10.7 & 10.7 & 11.5 & 11.5 \\
\midrule
\multicolumn{1}{l|}{\textbf{COMMONSENSE}} & & & & & & \\
story\_cloze-em & 78.0 & 96.0 & 92.0 & 78.0 & 94.0 & 90.0 \\
piqa-em        & 42.0 & 48.0 & 44.0 & 42.0 & 42.0 & 44.0 \\
copa-em        & 68.0 & 86.0 & 72.0 & 68.0 & 78.0 & 70.0 \\
hellaswag-em   & 34.0 & 44.0 & 30.0 & 36.0 & 42.0 & 22.0 \\
\midrule
\multicolumn{1}{l|}{\textbf{sentiment}} & & & & & & \\
sst2-em                   & 32.0 & 98.0 & 78.0 & 34.0 & 100.0 & 70.0 \\
yelp\_polarity\_reviews-em & 46.0 & 98.0 & 88.0 & 46.0 & 98.0 & 86.0 \\
imdb\_reviews-em           & 48.0 & 96.0 & 68.0 & 42.0 & 98.0 & 62.0 \\
sentiment140-em           & 46.0 & 70.0 & 66.0 & 44.0 & 68.0 & 72.0 \\
\midrule
\multicolumn{1}{l|}{\textbf{READING Comp.}} & & & & & & \\
multirc-em    & 38.0 & 48.0 & 54.0 & 34.0 & 42.0 & 46.0 \\
squad\_v2-em   & 34.0 & 26.0 & 26.0 & 34.0 & 24.0 & 30.0 \\
squad\_v1-em   & 60.0 & 68.0 & 62.0 & 60.0 & 68.0 & 48.0 \\
openbookqa-em & 68.0 & 76.0 & 68.0 & 68.0 & 72.0 & 70.0 \\
bool\_q-em     & 64.0 & 88.0 & 82.0 & 72.0 & 84.0 & 78.0 \\
drop-em       & 20.0 & 28.0 & 18.0 & 20.0 & 28.0 & 8.0 \\
\midrule
\multicolumn{1}{l|}{\textbf{CLOSE-BOOK QA}} & & & & & & \\
natural\_questions-em & 16.0 & 24.0 & 14.0 & 14.0 & 22.0 & 16.0 \\
arc\_easy-em          & 94.0 & 94.0 & 88.0 & 94.0 & 94.0 & 80.0 \\
arc\_challenge-em     & 58.0 & 70.0 & 54.0 & 62.0 & 64.0 & 52.0 \\
trivia\_qa-em         & 68.0 & 70.0 & 62.0 & 68.0 & 70.0 & 58.0 \\
\midrule
\multicolumn{1}{l|}{\textbf{COREFERENCE}} & & & & & & \\
definite\_pronoun\_resolution-em & 54.0 & 76.0 & 64.0 & 62.0 & 70.0 & 70.0 \\
wsc-em                           & 36.0 & 46.0 & 42.0 & 34.0 & 52.0 & 42.0 \\
\midrule
\multicolumn{1}{l|}{\textbf{READ. COMP. W/ COM}} & & & & & & \\
cosmos\_qa-em & 42.0 & 82.0 & 64.0 & 42.0 & 70.0 & 52.0 \\
record-em    & 18.0 & 36.0 & 14.0 & 18.0 & 34.0 & 14.0 \\
\midrule
\multicolumn{1}{l|}{\textbf{PARAPHRASE}} & & & & & & \\
paws\_wiki-em & 42.0 & 62.0 & 52.0 & 42.0 & 68.0 & 44.0 \\
glue\_qqp-em  & 42.0 & 58.0 & 56.0 & 34.0 & 68.0 & 42.0 \\
glue\_mrpc-em & 56.0 & 64.0 & 58.0 & 56.0 & 62.0 & 60.0 \\
stsb-em       & 14.0 & 26.0 & 12.0 & 14.0 & 20.0 & 4.0 \\
\midrule
\multicolumn{1}{l|}{\textbf{NLI}} & & & & & & \\
cb-em               & 48.9 & 82.2 & 75.6 & 44.4 & 86.7 & 57.8 \\
wnli-em             & 44.0 & 54.0 & 68.0 & 44.0 & 60.0 & 44.0 \\
anli\_r1-em          & 46.0 & 48.0 & 36.0 & 46.0 & 48.0 & 34.0 \\
anli\_r2-em          & 30.0 & 50.0 & 34.0 & 32.0 & 42.0 & 34.0 \\
anli\_r3-em          & 30.0 & 50.0 & 32.0 & 30.0 & 50.0 & 30.0 \\
mnli\_matched-em     & 60.0 & 92.0 & 86.0 & 64.0 & 92.0 & 86.0 \\
mnli\_mismatched-em  & 64.0 & 94.0 & 74.0 & 68.0 & 94.0 & 78.0 \\
snli-em              & 52.0 & 94.0 & 64.0 & 52.0 & 94.0 & 58.0 \\
qnli-em              & 50.0 & 70.0 & 58.0 & 54.0 & 72.0 & 58.0 \\
rte-em               & 60.0 & 68.0 & 70.0 & 62.0 & 68.0 & 72.0 \\
\bottomrule
\end{tabular}
}
\end{sc}
\end{small}
\end{center}
\end{table*}

\begin{table*}[t]
\caption{Per-task results of \sys for Llama2-7B with two large adapter pools.}
\vskip -0.1in
\label{tab:full_results_hf}
\begin{center}
\begin{small}
\begin{sc}
\resizebox{0.7\textwidth}{!}{%
\begin{tabular}{l|c c c|c c}
\toprule
& \multicolumn{3}{c|}{HF+48} & \multicolumn{2}{c}{HF} \\
\cmidrule(lr){2-4}\cmidrule(lr){5-6}
Domain-Metric & Non-OOD & Semi-OOD & OOD & Semi-OOD & OOD \\
\midrule
\multicolumn{1}{l|}{\textbf{Struct to Text}} & & & & & \\
web\_nlg\_en-rouge-1 & 63.6 & 52.7 & 50.4 & 32.3 & 46.8 \\
web\_nlg\_en-rouge-2 & 42.4 & 29.4 & 27.6 & 17.4 & 23.7 \\
web\_nlg\_en-rouge-l & 57.5 & 48.2 & 45.6 & 29.7 & 43.3 \\
dart-rouge-1         & 61.0 & 61.0 & 58.9 & 41.3 & 57.2 \\
dart-rouge-2         & 35.7 & 35.6 & 33.7 & 20.9 & 31.2 \\
dart-rouge-l         & 53.7 & 54.0 & 52.7 & 38.2 & 52.2 \\
e2e\_nlg-rouge-1      & 63.7 & 63.2 & 58.0 & 48.4 & 58.2 \\
e2e\_nlg-rouge-2      & 37.2 & 36.5 & 33.0 & 26.7 & 31.6 \\
e2e\_nlg-rouge-l      & 53.6 & 52.3 & 50.6 & 39.4 & 50.8 \\
common\_gen-rouge-1   & 47.0 & 28.1 & 31.0 & 22.9 & 32.5 \\
common\_gen-rouge-2   & 19.1 & 7.1 & 8.1 & 7.4 & 10.5 \\
common\_gen-rouge-l   & 43.0 & 24.3 & 26.6 & 20.6 & 27.6 \\
\midrule
\multicolumn{1}{l|}{\textbf{Translation}} & & & & & \\
para\_crawl\_enes-bleu      & 23.5 & 16.1 & 22.3 & 8.3 & 23.4 \\
wmt16\_translate\_tren-bleu & 2.0 & 2.0 & 3.2 & 1.0 & 2.3 \\
wmt16\_translate\_deen-bleu & 19.4 & 17.8 & 19.9 & 9.1 & 16.0 \\
wmt16\_translate\_ruen-bleu & 11.6 & 8.1 & 8.7 & 2.0 & 10.5 \\
wmt16\_translate\_fien-bleu & 6.2 & 7.2 & 6.9 & 2.5 & 5.0 \\
wmt16\_translate\_roen-bleu & 13.3 & 14.1 & 11.0 & 4.7 & 10.9 \\
wmt14\_enfr-bleu            & 16.0 & 16.2 & 17.9 & 9.8 & 17.1 \\
wmt16\_translate\_csen-bleu & 6.9 & 5.6 & 4.3 & 4.5 & 6.7 \\
\midrule
\multicolumn{1}{l|}{\textbf{COMMONSENSE}} & & & & & \\
story\_cloze-em & 96.0 & 90.0 & 88.0 & 94.0 & 92.0 \\
piqa-em        & 44.0 & 40.0 & 40.0 & 38.0 & 38.0 \\
copa-em        & 82.0 & 82.0 & 82.0 & 80.0 & 80.0 \\
hellaswag-em   & 50.0 & 50.0 & 46.0 & 42.0 & 52.0 \\
\midrule
\multicolumn{1}{l|}{\textbf{sentiment}} & & & & & \\
sst2-em                   & 98.0 & 98.0 & 98.0 & 94.0 & 92.0 \\
yelp\_polarity\_reviews-em & 98.0 & 98.0 & 98.0 & 98.0 & 96.0 \\
imdb\_reviews-em           & 98.0 & 98.0 & 100.0 & 98.0 & 100.0 \\
sentiment140-em           & 68.0 & 68.0 & 70.0 & 64.0 & 68.0 \\
\midrule
\multicolumn{1}{l|}{\textbf{READING Comp.}} & & & & & \\
multirc-em    & 62.0 & 56.0 & 42.0 & 54.0 & 50.0 \\
squad\_v2-em   & 54.0 & 20.0 & 18.0 & 32.0 & 32.0 \\
squad\_v1-em   & 74.0 & 72.0 & 66.0 & 54.0 & 66.0 \\
openbookqa-em & 80.0 & 78.0 & 76.0 & 66.0 & 66.0 \\
bool\_q-em     & 78.0 & 80.0 & 72.0 & 82.0 & 76.0 \\
drop-em       & 22.0 & 14.0 & 16.0 & 10.0 & 16.0 \\
\midrule
\multicolumn{1}{l|}{\textbf{CLOSE-BOOK QA}} & & & & & \\
natural\_questions-em & 16.0 & 12.0 & 12.0 & 4.0 & 10.0 \\
arc\_easy-em          & 78.0 & 78.0 & 84.0 & 70.0 & 70.0 \\
arc\_challenge-em     & 50.0 & 48.0 & 46.0 & 38.0 & 38.0 \\
trivia\_qa-em         & 50.0 & 46.0 & 46.0 & 36.0 & 48.0 \\
\midrule
\multicolumn{1}{l|}{\textbf{COREFERENCE}} & & & & & \\
definite\_pronoun\_resolution-em & 62.0 & 64.0 & 62.0 & 70.0 & 66.0 \\
wsc-em                           & 58.0 & 54.0 & 46.0 & 58.0 & 58.0 \\
\midrule
\multicolumn{1}{l|}{\textbf{READ. COMP. W/ COM}} & & & & & \\
cosmos\_qa-em & 72.0 & 54.0 & 62.0 & 52.0 & 46.0 \\
record-em    & 64.0 & 34.0 & 30.0 & 22.0 & 22.0 \\
\midrule
\multicolumn{1}{l|}{\textbf{PARAPHRASE}} & & & & & \\
paws\_wiki-em & 60.0 & 66.0 & 52.0 & 66.0 & 54.0 \\
glue\_qqp-em  & 80.0 & 78.0 & 76.0 & 82.0 & 52.0 \\
glue\_mrpc-em & 66.0 & 66.0 & 62.0 & 50.0 & 34.0 \\
stsb-em       & 38.0 & 18.0 & 20.0 & 12.0 & 14.0 \\
\midrule
\multicolumn{1}{l|}{\textbf{NLI}} & & & & & \\
cb-em               & 88.9 & 84.4 & 84.4 & 86.7 & 84.4 \\
wnli-em             & 52.0 & 52.0 & 54.0 & 62.0 & 64.0 \\
anli\_r1-em          & 54.0 & 50.0 & 46.0 & 52.0 & 52.0 \\
anli\_r2-em          & 44.0 & 40.0 & 40.0 & 40.0 & 38.0 \\
anli\_r3-em          & 44.0 & 54.0 & 56.0 & 54.0 & 48.0 \\
mnli\_matched-em     & 88.0 & 86.0 & 86.0 & 82.0 & 90.0 \\
mnli\_mismatched-em  & 94.0 & 94.0 & 92.0 & 98.0 & 98.0 \\
snli-em              & 84.0 & 84.0 & 84.0 & 90.0 & 88.0 \\
qnli-em              & 66.0 & 72.0 & 72.0 & 74.0 & 70.0 \\
rte-em               & 84.0 & 66.0 & 66.0 & 64.0 & 64.0 \\
\bottomrule
\end{tabular}
}
\end{sc}
\end{small}
\end{center}
\end{table*}

\begin{table*}[t]
\caption{Per-task results for selection-based adapter routing methods on LLaMA2-7B, including oracle, performance-aligned, and input-aware selection. Results for \loraretriever-based selection are directly copied from the original \loraretriever paper.}
\label{tab:full_results_selection_7b}
\begin{center}
\begin{small}
\begin{sc}
\resizebox{0.9\textwidth}{!}{%
\begin{tabular}{
l|c|c|c|c|c|c|c}
\toprule
\multicolumn{8}{c}{\textbf{Llama2-7B}} \\
\midrule
 & \multicolumn{2}{c|} {Oracle}
 & \multicolumn{2}{c|}{Non-OOD}
 & \multicolumn{1}{c|}{Semi-OOD}
 & \multicolumn{2}{c}{OOD}\\
\cmidrule(lr){2-8}
Domain-Metric
& \makecell{Task-\\Aligned}
& \makecell{Perf-\\Aligned}
& \makecell{LoRA-\\Retriever\\Selection}
& \makecell{LoRAuter\\Selection}
& \makecell{LoRAuter\\Selection}
& \makecell{LoRA-\\Retriever\\Selection}
& \makecell{LoRAuter\\Selection} \\
\midrule
\multicolumn{1}{l|}{\textbf{Struct to Text}} & & & & & & & \\
web\_nlg\_en-rouge-1 & 69.6 & 69.6 & 67.0 & 66.2 & 54.5 & 53.9 & 51.4 \\
web\_nlg\_en-rouge-2 & 48.0 & 48.0 & 44.5 & 44.8 & 30.0 & 30.0 & 27.8 \\
web\_nlg\_en-rouge-l & 62.8 & 62.8 & 60.9 & 59.9 & 49.2 & 49.1 & 46.7 \\
dart-rouge-1       & 71.7 & 71.7 & 67.9 & 67.6 & 60.2 & 58.4 & 51.3 \\
dart-rouge-2       & 49.4 & 49.4 & 45.8 & 46.3 & 35.2 & 34.9 & 27.9 \\
dart-rouge-l       & 64.9 & 64.9 & 61.1 & 61.8 & 53.5 & 52.4 & 46.3 \\
e2e\_nlg-rouge-1    & 66.1 & 66.1 & 65.8 & 63.9 & 52.2 & 59.3 & 41.5 \\
e2e\_nlg-rouge-2    & 39.6 & 39.6 & 39.4 & 37.4 & 27.6 & 34.1 & 19.6 \\
e2e\_nlg-rouge-l    & 56.4 & 56.4 & 55.7 & 54.2 & 46.2 & 50.2 & 36.5 \\
common\_gen-rouge-1 & 46.5 & 46.5 & 44.7 & 46.2 & 29.3 & 29.0 & 29.5 \\
common\_gen-rouge-2 & 18.9 & 18.9 & 18.3 & 18.9 & 7.5  & 7.3 & 7.8 \\
common\_gen-rouge-l & 42.2 & 42.2 & 40.5 & 42.4 & 24.0 & 24.0 & 24.4 \\
\midrule
\multicolumn{1}{l|}{\textbf{Translation}}  & & & & & & & \\
para\_crawl\_enes-bleu        & 24.0 & 24.0 & 24.2 & 23.8 & 19.3 & 20.3 & 19.6 \\
wmt16\_translate\_tren-bleu   & 3.7  & 2.0  & 3.1 & 2.0  & 2.0  & 2.6 & 1.9 \\
wmt16\_translate\_deen-bleu   & 17.8 & 17.8 & 10.4 & 17.8 & 16.9 & 9.8 & 16.3 \\
wmt16\_translate\_ruen-bleu   & 11.5 & 11.5 & 18.7 & 11.5 & 11.9 & 20.3 & 11.3 \\
wmt16\_translate\_fien-bleu   & 6.8  & 8.0  & 6.5 & 8.0  & 8.0  & 7.0 & 7.4 \\
wmt16\_translate\_roen-bleu   & 14.6 & 13.5 & 14.0 & 13.5 & 13.5 & 12.3 & 11.9 \\
wmt14\_enfr-bleu              & 16.0 & 17.0 & 16.1 & 17.0 & 17.0 & 16.9 & 18.0 \\
wmt16\_translate\_csen-bleu   & 10.5 & 6.9  & 9.4 & 6.9  & 6.9  & 7.0 & 7.0 \\
\midrule
\multicolumn{1}{l|}{\textbf{COMMONSENSE}} & & & & & & & \\
story\_cloze-em & 72.0 & 96.0 & 62.0 & 96.0 & 96.0 & 42.0 & 84.0 \\
piqa-em        & 44.0 & 44.0 & 46.0 & 44.0 & 34.0 & 32.0 & 36.0 \\
copa-em        & 86.0 & 84.0 & 74.0 & 84.0 & 84.0 & 68.0 & 78.0 \\
hellaswag-em   & 50.0 & 50.0 & 40.0 & 50.0 & 48.0 & 42.0 & 50.0 \\
\midrule
\multicolumn{1}{l|}{\textbf{sentiment}} & & & & & & & \\
sst2-em                   & 98.0 & 98.0 & 98.0 & 98.0 & 96.0 & 96.0 & 96.0 \\
yelp\_polarity\_reviews-em & 98.0 & 98.0 & 94.0 & 98.0 & 98.0 & 94.0 & 100.0 \\
imdb\_reviews-em           & 96.0 & 98.0 & 96.0 & 96.0 & 94.0 & 96.0 & 96.0 \\
sentiment140-em           & 68.0 & 68.0 & 70.0 & 68.0 & 62.0 & 70.0 & 68.0 \\
\midrule
\multicolumn{1}{l|}{\textbf{READING Comp.}} & & & & & & & \\
multirc-em      & 68.0 & 68.0 & 52.0 & 68.0 & 66.0 & 38.0 & 62.0 \\
squad\_v2-em     & 62.0 & 62.0 & 56.0 & 62.0 & 20.0 & 12.0 & 14.0 \\
squad\_v1-em     & 68.0 & 68.0 & 66.0 & 68.0 & 72.0 & 68.0 & 62.0 \\
openbookqa-em   & 84.0 & 84.0 & 68.0 & 76.0 & 66.0 & 58.0 & 76.0 \\
bool\_q-em       & 84.0 & 84.0 & 60.0 & 80.0 & 78.0 & 60.0 & 72.0 \\
drop-em         & 40.0 & 40.0 & 8.0 & 28.0 & 20.0 & 6.0 & 6.0 \\
\midrule
\multicolumn{1}{l|}{\textbf{CLOSE-BOOK QA}} & & & & & & & \\
natural\_questions-em & 18.0 & 18.0 & 16.0 & 16.0 & 10.0 & 10.0 & 10.0 \\
arc\_easy-em          & 48.0 & 74.0 & 56.0 & 74.0 & 66.0 & 70.0 & 74.0 \\
arc\_challenge-em     & 46.0 & 46.0 & 42.0 & 46.0 & 40.0 & 46.0 & 46.0 \\
trivia\_qa-em         & 64.0 & 64.0 & 46.0 & 44.0 & 44.0 & 46.0 & 34.0 \\
\midrule
\multicolumn{1}{l|}{\textbf{COREFERENCE}}  & & & & & & & \\
definite\_pronoun\_resolution-em & 52.0 & 64.0 & 50.0 & 64.0 & 64.0 & 50.0 & 50.0 \\
wsc-em                             & 50.0 & 54.0 & 50.0 & 54.0 & 54.0 & 42.0 & 54.0 \\
\midrule
\multicolumn{1}{l|}{\textbf{READ. COMP. W/ COM}}  & & & & & & & \\
cosmos\_qa-em & 70.0 & 70.0 & 68.0 & 64.0 & 50.0 & 34.0 & 18.0 \\
record-em    & 70.0 & 70.0 & 70.0 & 54.0 & 22.0 & 26.0 & 20.0 \\
\midrule
\multicolumn{1}{l|}{\textbf{PARAPHRASE}} & & & & & & & \\
paws\_wiki-em   & 90.0 & 90.0 & 64.0 & 60.0 & 52.0 & 40.0 & 50.0 \\
glue\_qqp-em    & 70.0 & 82.0 & 74.0 & 82.0 & 82.0 & 68.0 & 78.0 \\
glue\_mrpc-em   & 60.0 & 70.0 & 58.0 & 66.0 & 72.0 & 58.0 & 58.0 \\
stsb-em        & 34.0 & 34.0 & 36.0 & 34.0 & 18.0 & 16.0 & 26.0 \\
\midrule
\multicolumn{1}{l|}{\textbf{NLI}} & & & & & & & \\
cb-em               & 88.9 & 88.9 & 80.0 & 86.7 & 75.6 & 62.2 & 57.8 \\
wnli-em             & 72.0 & 64.0 & 68.0 & 64.0 & 64.0 & 46.0 & 46.0 \\
anli\_r1-em          & 52.0 & 52.0 & 50.0 & 48.0 & 46.0 & 50.0 & 44.0 \\
anli\_r2-em          & 46.0 & 38.0 & 46.0 & 42.0 & 38.0 & 46.0 & 46.0 \\
anli\_r3-em          & 46.0 & 38.0 & 42.0 & 44.0 & 44.0 & 38.0 & 54.0 \\
mnli\_matched-em     & 88.0 & 88.0 & 84.0 & 88.0 & 88.0 & 88.0 & 88.0 \\
mnli\_mismatched-em  & 92.0 & 92.0 & 90.0 & 92.0 & 84.0 & 94.0 & 94.0 \\
snli-em             & 96.0 & 84.0 & 84.0 & 84.0 & 82.0 & 84.0 & 82.0 \\
qnli-em             & 94.0 & 94.0 & 94.0 & 88.0 & 68.0 & 26.0 & 64.0 \\
rte-em              & 56.0 & 86.0 & 62.0 & 80.0 & 86.0 & 72.0 & 64.0 \\
\bottomrule
\end{tabular}
}
\end{sc}
\end{small}
\end{center}
\vskip -0.1in
\end{table*}

\begin{table*}[t]
\caption{Per-task results for selection-based adapter routing methods on LLaMA2-13B, including oracle, performance-aligned, and input-aware selection. Results for \loraretriever-based selection are directly copied from the original \loraretriever paper.}
\label{tab:full_results_selection_13b}
\begin{center}
\begin{small}
\begin{sc}
\resizebox{0.9\textwidth}{!}{%
\begin{tabular}{l|c|c|c|c|c|c|c}
\toprule
\multicolumn{8}{c}{\textbf{Llama2-7B}} \\
\midrule
 & \multicolumn{2}{c|} {Oracle}
 & \multicolumn{2}{c|}{Non-OOD}
 & \multicolumn{1}{c|}{Semi-OOD}
 & \multicolumn{2}{c}{OOD}\\
\cmidrule(lr){2-8}
Domain-Metric
& \makecell{Task-\\Aligned}
& \makecell{Perf-\\Aligned}
& \makecell{LoRA-\\Retriever\\Selection}
& \makecell{LoRAuter\\Selection}
& \makecell{LoRAuter\\Selection}
& \makecell{LoRA-\\Retriever\\Selection}
& \makecell{LoRAuter\\Selection} \\
\midrule
\multicolumn{1}{l|}{\textbf{Struct to Text}} & & & & & & & \\
web\_nlg\_en-rouge-1 & 72.4 & 72.4 & 68.5 & 68.1 & 56.7 & 51.9 & 51.1 \\
web\_nlg\_en-rouge-2 & 52.1 & 52.1 & 47.5 & 48.8 & 33.0 & 28.6 & 29.4 \\
web\_nlg\_en-rouge-l & 66.1 & 66.1 & 62.4 & 62.2 & 50.8 & 48.3 & 46.4 \\
dart-rouge-1       & 72.9 & 72.9 & 67.0 & 69.4 & 59.7 & 57.0 & 51.3 \\
dart-rouge-2       & 52.9 & 52.9 & 45.9 & 49.5 & 35.3 & 33.6 & 29.2 \\
dart-rouge-l       & 66.7 & 66.7 & 61.2 & 62.9 & 53.5 & 50.0 & 45.6 \\
e2e\_nlg-rouge-1    & 66.5 & 66.5 & 66.1 & 64.3 & 57.8 & 59.2 & 44.4 \\
e2e\_nlg-rouge-2    & 38.8 & 38.8 & 39.3 & 36.6 & 31.0 & 32.8 & 20.5 \\
e2e\_nlg-rouge-l    & 56.4 & 56.4 & 56.4 & 54.3 & 45.9 & 48.9 & 37.9 \\
common\_gen-rouge-1 & 48.4 & 48.4 & 48.9 & 48.6 & 27.6 & 29.3 & 29.5 \\
common\_gen-rouge-2 & 17.9 & 17.9 & 20.3 & 17.9 & 7.8  & 8.2 & 8.6 \\
common\_gen-rouge-l & 44.1 & 44.1 & 44.1 & 44.5 & 25.7 & 24.4 & 25.5 \\
\midrule
\multicolumn{1}{l|}{\textbf{Translation}}  & & & & & & & \\
para\_crawl\_enes-bleu        & 25.0 & 25.0 & 25.4 & 24.9 & 22.1 & 23.1 & 23.1 \\
wmt16\_translate\_tren-bleu   & 2.0  & 3.2  & 2.4 & 3.2  & 3.2  & 1.2 & 2.5 \\
wmt16\_translate\_deen-bleu   & 19.8 & 19.8 & 11.5 & 19.8 & 20.7 & 10.3 & 18.0 \\
wmt16\_translate\_ruen-bleu   & 11.3 & 10.9 & 19.9 & 10.9 & 10.9 & 20.7 & 11.0 \\
wmt16\_translate\_fien-bleu   & 6.8  & 7.3  & 6.8 & 7.3  & 7.3  & 5.1 & 6.6 \\
wmt16\_translate\_roen-bleu   & 14.2 & 10.2 & 13.9 & 10.2 & 10.2 & 10.9 & 12.9 \\
wmt14\_enfr-bleu              & 16.5 & 20.4 & 17.1 & 20.4 & 20.4 & 18.4 & 18.8 \\
wmt16\_translate\_csen-bleu   & 6.9  & 8.5  & 6.2 & 8.5  & 8.5  & 11.6 & 9.2 \\
\midrule
\multicolumn{1}{l|}{\textbf{COMMONSENSE}} & & & & & & & \\
story\_cloze-em & 94.0 & 94.0 & 80.0 & 90.0 & 90.0 & 56.0 & 96.0 \\
piqa-em        & 48.0 & 44.0 & 52.0 & 44.0 & 44.0 & 30.0 & 40.0 \\
copa-em        & 76.0 & 78.0 & 74.0 & 78.0 & 78.0 & 68.0 & 78.0 \\
hellaswag-em   & 58.0 & 58.0 & 30.0 & 52.0 & 44.0 & 36.0 & 48.0 \\
\midrule
\multicolumn{1}{l|}{\textbf{sentiment}} & & & & & & & \\
sst2-em                   & 98.0 & 98.0 & 98.0 & 98.0 & 98.0 & 98.0 & 100.0 \\
yelp\_polarity\_reviews-em & 98.0 & 98.0 & 98.0 & 98.0 & 98.0 & 100.0 & 100.0 \\
imdb\_reviews-em           & 96.0 & 96.0 & 98.0 & 98.0 & 98.0 & 98.0 & 98.0 \\
sentiment140-em           & 68.0 & 68.0 & 68.0 & 66.0 & 66.0 & 68.0 & 70.0 \\
\midrule
\multicolumn{1}{l|}{\textbf{READING Comp.}} & & & & & & & \\
multirc-em      & 88.0 & 88.0 & 72.0 & 86.0 & 74.0 & 36.0 & 40.0 \\
squad\_v2-em     & 68.0 & 68.0 & 62.0 & 68.0 & 40.0 & 62.0 & 14.0 \\
squad\_v1-em     & 74.0 & 74.0 & 58.0 & 68.0 & 70.0 & 34.0 & 52.0 \\
openbookqa-em   & 86.0 & 86.0 & 80.0 & 78.0 & 74.0 & 72.0 & 70.0 \\
bool\_q-em       & 84.0 & 84.0 & 68.0 & 82.0 & 76.0 & 66.0 & 80.0 \\
drop-em         & 58.0 & 58.0 & 22.0 & 52.0 & 30.0 & 18.0 & 20.0 \\
\midrule
\multicolumn{1}{l|}{\textbf{CLOSE-BOOK QA}} & & & & & & & \\
natural\_questions-em & 30.0 & 30.0 & 28.0 & 24.0 & 18.0 & 12.0 & 14.0 \\
arc\_easy-em          & 94.0 & 94.0 & 90.0 & 92.0 & 84.0 & 88.0 & 86.0 \\
arc\_challenge-em     & 68.0 & 62.0 & 66.0 & 70.0 & 66.0 & 58.0 & 68.0 \\
trivia\_qa-em         & 68.0 & 70.0 & 56.0 & 58.0 & 64.0 & 54.0 & 58.0 \\
\midrule
\multicolumn{1}{l|}{\textbf{COREFERENCE}}  & & & & & & & \\
definite\_pronoun\_resolution-em & 88.0 & 88.0 & 90.0 & 88.0 & 74.0 & 72.0 & 62.0 \\
wsc-em                             & 62.0 & 64.0 & 60.0 & 64.0 & 64.0 & 58.0 & 48.0 \\
\midrule
\multicolumn{1}{l|}{\textbf{READ. COMP. W/ COM}}  & & & & & & & \\
cosmos\_qa-em & 84.0 & 84.0 & 82.0 & 74.0 & 74.0 & 38.0 & 30.0 \\
record-em    & 78.0 & 78.0 & 78.0 & 60.0 & 34.0 & 28.0 & 38.0 \\
\midrule
\multicolumn{1}{l|}{\textbf{PARAPHRASE}} & & & & & & & \\
paws\_wiki-em   & 92.0 & 92.0 & 70.0 & 62.0 & 56.0 & 52.0 & 50.0 \\
glue\_qqp-em    & 86.0 & 86.0 & 88.0 & 82.0 & 84.0 & 76.0 & 80.0 \\
glue\_mrpc-em   & 84.0 & 84.0 & 70.0 & 80.0 & 70.0 & 62.0 & 62.0 \\
stsb-em         & 44.0 & 44.0 & 44.0 & 44.0 & 28.0 & 20.0 & 22.0 \\
\midrule
\multicolumn{1}{l|}{\textbf{NLI}} & & & & & & & \\
cb-em               & 95.6 & 95.6 & 91.1 & 95.6 & 91.1 & 84.4 & 91.1 \\
wnli-em             & 72.0 & 72.0 & 72.0 & 72.0 & 58.0 & 62.0 & 52.0 \\
anli\_r1-em          & 70.0 & 74.0 & 70.0 & 70.0 & 54.0 & 70.0 & 70.0 \\
anli\_r2-em          & 66.0 & 66.0 & 56.0 & 66.0 & 38.0 & 56.0 & 64.0 \\
anli\_r3-em          & 68.0 & 68.0 & 56.0 & 60.0 & 64.0 & 56.0 & 60.0 \\
mnli\_matched-em     & 86.0 & 88.0 & 90.0 & 88.0 & 88.0 & 88.0 & 88.0 \\
mnli\_mismatched-em  & 90.0 & 90.0 & 90.0 & 90.0 & 88.0 & 94.0 & 90.0 \\
snli-em              & 90.0 & 90.0 & 88.0 & 90.0 & 82.0 & 88.0 & 88.0 \\
qnli-em              & 94.0 & 94.0 & 94.0 & 84.0 & 66.0 & 30.0 & 58.0 \\
rte-em               & 88.0 & 88.0 & 82.0 & 80.0 & 66.0 & 74.0 & 74.0 \\
\bottomrule
\end{tabular}
}
\end{sc}
\end{small}
\end{center}
\vskip -0.1in
\end{table*}

\begin{table*}[t]
\caption{Per-task results for the \loraretriever weighted mixture routing method on LLaMA2-7B and LLaMA2-13B under Non-OOD and OOD evaluation.}
\label{tab:full_results_weighted_mixture}
\begin{center}
\begin{small}
\begin{sc}
\resizebox{0.78\textwidth}{!}{%
\begin{tabular}{
l|
C{2.2cm}|C{2.2cm}|
C{2.2cm}|C{2.2cm}
}
\toprule
 & \multicolumn{2}{c|}{\textbf{Llama2-7B}}
 & \multicolumn{2}{c}{\textbf{Llama2-13B}} \\
\cmidrule(lr){2-3} \cmidrule(lr){4-5}
Domain-Metric
& \makecell{Non-OOD}
& \makecell{OOD}
& \makecell{Non-OOD}
& \makecell{OOD} \\
\midrule
\multicolumn{1}{l|}{\textbf{Struct to Text}} & & & & \\
web\_nlg\_en-rouge-1 & 60.9 & 48.5 & 68.6 & 52.2 \\
web\_nlg\_en-rouge-2 & 36.7 & 26.5 & 47.2 & 28.5 \\
web\_nlg\_en-rouge-l & 54.0 & 43.8 & 61.6 & 46.7 \\
dart-rouge-1       & 62.7 & 58.0 & 64.1 & 59.2 \\
dart-rouge-2       & 36.7 & 33.4 & 39.9 & 36.1 \\
dart-rouge-l       & 55.9 & 52.3 & 55.7 & 52.2 \\
e2e\_nlg-rouge-1    & 64.3 & 55.5 & 64.2 & 61.8 \\
e2e\_nlg-rouge-2    & 37.8 & 29.8 & 36.6 & 34.4 \\
e2e\_nlg-rouge-l    & 54.4 & 47.9 & 54.4 & 52.1 \\
common\_gen-rouge-1 & 50.1 & 29.8 & 46.1 & 28.2 \\
common\_gen-rouge-2 & 23.3 & 9.1  & 21.2 & 9.6 \\
common\_gen-rouge-l & 46.5 & 26.2 & 42.6 & 24.4 \\
\midrule
\multicolumn{1}{l|}{\textbf{Translation}} & & & & \\
para\_crawl\_enes-bleu        & 23.4 & 23.0 & 25.7 & 23.6 \\
wmt16\_translate\_tren-bleu   & 3.2  & 2.6  & 2.5  & 2.8 \\
wmt16\_translate\_deen-bleu   & 20.0 & 19.5 & 20.8 & 20.8 \\
wmt16\_translate\_ruen-bleu   & 10.8 & 8.5  & 10.3 & 11.0 \\
wmt16\_translate\_fien-bleu   & 6.5  & 7.1  & 6.8  & 8.3 \\
wmt16\_translate\_roen-bleu   & 14.6 & 13.3 & 15.8 & 16.2 \\
wmt14\_enfr-bleu              & 17.3 & 18.2 & 15.7 & 18.8 \\
wmt16\_translate\_csen-bleu   & 10.0 & 5.8  & 9.4  & 9.9 \\
\midrule
\multicolumn{1}{l|}{\textbf{COMMONSENSE}} & & & & \\
story\_cloze-em & 80.0 & 88.0 & 96.0 & 94.0 \\
piqa-em        & 42.0 & 36.0 & 52.0 & 48.0 \\
copa-em        & 84.0 & 78.0 & 80.0 & 80.0 \\
hellaswag-em   & 34.0 & 30.0 & 52.0 & 48.0 \\
\midrule
\multicolumn{1}{l|}{\textbf{sentiment}} & & & & \\
sst2-em                   & 98.0 & 94.0 & 98.0 & 100.0 \\
yelp\_polarity\_reviews-em & 98.0 & 98.0 & 100.0 & 100.0 \\
imdb\_reviews-em           & 98.0 & 100.0 & 98.0 & 98.0 \\
sentiment140-em           & 68.0 & 72.0 & 72.0 & 72.0 \\
\midrule
\multicolumn{1}{l|}{\textbf{READING Comp.}} & & & & \\
multirc-em      & 62.0 & 46.0 & 78.0 & 40.0 \\
squad\_v2-em     & 56.0 & 14.0 & 66.0 & 20.0 \\
squad\_v1-em     & 64.0 & 64.0 & 68.0 & 62.0 \\
openbookqa-em   & 74.0 & 70.0 & 88.0 & 80.0 \\
bool\_q-em       & 80.0 & 78.0 & 84.0 & 86.0 \\
drop-em         & 20.0 & 14.0 & 38.0 & 30.0 \\
\midrule
\multicolumn{1}{l|}{\textbf{CLOSE-BOOK QA}}  & & & & \\
natural\_questions-em & 12.0 & 10.0 & 28.0 & 18.0 \\
arc\_easy-em          & 74.0 & 80.0 & 94.0 & 90.0 \\
arc\_challenge-em     & 46.0 & 46.0 & 80.0 & 72.0 \\
trivia\_qa-em         & 56.0 & 50.0 & 62.0 & 56.0 \\
\midrule
\multicolumn{1}{l|}{\textbf{COREFERENCE}}  & & & & \\
definite\_pronoun\_resolution-em & 60.0 & 64.0 & 92.0 & 68.0 \\
wsc-em                             & 56.0 & 50.0 & 70.0 & 56.0 \\
\midrule
\multicolumn{1}{l|}{\textbf{READ. COMP. W/ COM}}  & & & & \\
cosmos\_qa-em & 74.0 & 48.0 & 82.0 & 62.0 \\
record-em    & 58.0 & 24.0 & 76.0 & 50.0 \\
\midrule
\multicolumn{1}{l|}{\textbf{PARAPHRASE}}  & & & & \\
paws\_wiki-em   & 52.0 & 46.0 & 88.0 & 62.0 \\
glue\_qqp-em    & 86.0 & 60.0 & 82.0 & 82.0 \\
glue\_mrpc-em   & 62.0 & 64.0 & 72.0 & 64.0 \\
stsb-em        & 50.0 & 12.0 & 44.0 & 24.0 \\
\midrule
\multicolumn{1}{l|}{\textbf{NLI}} & & & & \\
cb-em               & 82.2 & 66.7 & 97.8 & 86.7 \\
wnli-em             & 66.0 & 44.0 & 74.0 & 64.0 \\
anli\_r1-em          & 52.0 & 44.0 & 72.0 & 68.0 \\
anli\_r2-em          & 48.0 & 48.0 & 60.0 & 64.0 \\
anli\_r3-em          & 48.0 & 46.0 & 64.0 & 60.0 \\
mnli\_matched-em     & 84.0 & 80.0 & 86.0 & 84.0 \\
mnli\_mismatched-em  & 90.0 & 84.0 & 94.0 & 92.0 \\
snli-em             & 86.0 & 84.0 & 92.0 & 88.0 \\
qnli-em             & 80.0 & 74.0 & 80.0 & 72.0 \\
rte-em              & 74.0 & 68.0 & 80.0 & 70.0 \\
\bottomrule
\end{tabular}
}
\end{sc}
\end{small}
\end{center}
\vskip -0.1in
\end{table*}

\newpage

\end{document}